\documentclass[11pt]{article}

\usepackage[margin=1in]{geometry}
\usepackage{iftex}
\ifPDFTeX
  \usepackage[utf8]{inputenc}
  \usepackage[T1]{fontenc}
  \usepackage{textcomp}
  \IfFileExists{lmodern.sty}{\usepackage{lmodern}\usepackage{microtype}}{}
  \DeclareUnicodeCharacter{0394}{\ensuremath{\Delta}}
  \DeclareUnicodeCharacter{03BA}{\ensuremath{\kappa}}
  \DeclareUnicodeCharacter{03C1}{\ensuremath{\rho}}
  \DeclareUnicodeCharacter{00B2}{\ensuremath{^2}}
  \DeclareUnicodeCharacter{00B3}{\ensuremath{^3}}
  \DeclareUnicodeCharacter{00B9}{\ensuremath{^1}}
  \DeclareUnicodeCharacter{02B0}{\ensuremath{^{\mathrm{h}}}}
  \DeclareUnicodeCharacter{0305}{}
  \DeclareUnicodeCharacter{2080}{\ensuremath{_0}}
  \DeclareUnicodeCharacter{2081}{\ensuremath{_1}}
  \DeclareUnicodeCharacter{2082}{\ensuremath{_2}}
  \DeclareUnicodeCharacter{2083}{\ensuremath{_3}}
  \DeclareUnicodeCharacter{2084}{\ensuremath{_4}}
  \DeclareUnicodeCharacter{207B}{\ensuremath{^-}}
  \DeclareUnicodeCharacter{2212}{\ensuremath{-}}
  \DeclareUnicodeCharacter{2002}{\enspace}
  \DeclareUnicodeCharacter{2009}{\,}
  \DeclareUnicodeCharacter{1D62}{\ensuremath{_{i}}}
  \DeclareUnicodeCharacter{1D63}{\ensuremath{_{r}}}
  \DeclareUnicodeCharacter{209C}{\ensuremath{_{t}}}
  \DeclareUnicodeCharacter{2C7C}{\ensuremath{_{j}}}
\else
  \usepackage{fontspec}
  \usepackage{unicode-math}
  \usepackage{microtype}
\fi

\usepackage{graphicx}
\graphicspath{{media/}}
\usepackage{longtable,booktabs,array,tabularx}
\usepackage{multirow}
\usepackage{calc}        
\usepackage{etoolbox}
\makeatletter
\patchcmd\longtable{\par}{\if@noskipsec\mbox{}\fi\par}{}{}
\makeatother
\IfFileExists{footnotehyper.sty}{\usepackage{footnotehyper}}{\usepackage{footnote}}
\makesavenoteenv{longtable}
\usepackage{caption}
\usepackage{amsmath}
\ifPDFTeX\usepackage{amssymb}\fi
\usepackage{siunitx}
\usepackage{xcolor}
\usepackage[round,authoryear]{natbib}
\usepackage[colorlinks=true,allcolors=blue,breaklinks=true]{hyperref}

\providecommand{\tightlist}{\setlength{\itemsep}{0pt}\setlength{\parskip}{0pt}}
\title{\bfseries Fusing Earth observation embeddings and spaceborne terrain with
meteorological forcing for flood peak depth prediction from high-water marks}

\author{
Wen Zhou\,$^{1,2}$ \quad
Shaowen Wang\,$^{1,2,3,}$\thanks{Corresponding author: \texttt{shaowen@illinois.edu}} \quad
Upmanu Lall\,$^{4}$ \quad
Venkatesh Merwade\,$^{5}$ \quad
Dingqi Ye\,$^{1,2}$ \\[8pt]
\begin{minipage}{0.95\textwidth}\centering\small
$^{1}$Department of Geography and Geographic Information Science,
$^{2}$CyberGIS Center for Advanced Digital and Spatial Studies, and
$^{3}$Siebel School of Computing and Data Science,
University of Illinois Urbana--Champaign, Urbana, IL, USA \\[2pt]
$^{4}$School of Complex Adaptive Systems, College of Global Futures,
Arizona State University, Tempe, AZ, USA \\[2pt]
$^{5}$Lyles School of Civil and Construction Engineering,
Purdue University, West Lafayette, IN, USA
\end{minipage}
}
\date{}

\begin{document}
\maketitle

\begin{abstract}
Flood peak depth on normally dry land governs damage and risk to life,
but is observed only sparsely, by post-event high-water mark (HWM)
surveys, and depth derived from satellite flood extent is produced after
the event and inherits the vertical error of the digital elevation model
(DEM) it is computed from. Deep-learning flood models have in turn been
trained on hydrodynamic simulations rather than on observed depths. We
ask whether flood peak depth can be learned from field observations and
predicted before the peak from spaceborne terrain, geospatial
foundation-model embeddings and antecedent meteorological forcing. We
assembled 5689 USGS HWM surveys from 39 events (2011-2025) in 24 U.S.
states and Puerto Rico, assigned each a peak time from nearby sensors,
paired each with a rain-free window at the same site, and linked them to
Copernicus GLO-30 terrain, AlphaEarth embeddings and 144 h of hourly
forcing masked to the lead time. A cross-modal fusion network trained on
these observations predicts depth at the surveyed marks with a mean
absolute error of 0.39 m 6 h ahead of the peak. Supervision by surveyed
depth corrects the DEM error implicitly: differencing the measured peak
water level with the same DEM gives 0.93 m at the same marks, 2.4 times
larger, although it uses the measured water level and the model does
not. Forcing accounts for 77\% of the attributed reduction in depth
error; terrain and the embeddings are statistically equivalent to each
other and contribute the remaining 23\%. On eleven held-out events the
model still separates flooded from rain-free conditions but no longer
resolves depth at the marks (R2 = 0.03), so the skill reported here is
confined to the events and regions sampled. Field-surveyed depths can
thus supervise flood depth prediction and correct terrain error that
extent-based estimation cannot, while transfer to unseen storms remains
the open problem.
\end{abstract}

\paragraph*{Highlights}
\begin{itemize}\tightlist
  \item Flood depth is learned from field-surveyed flood records, not from simulation
  \item Flood peak depth is forecast using only data available before the peak
  \item Model depth beats DEM subtraction even when the water level is measured
  \item AlphaEarth embeddings and the DEM are statistically equivalent for flood depth
  \item Depth skill does not transfer to unseen storms, which remains the open problem
\end{itemize}


\hypertarget{introduction}{%
\section{Introduction}\label{introduction}}

Floods are the most frequent weather-related hazard worldwide,
accounting for 35--40\% of weather-related disasters with recorded
events rising 134\% since 2000 \citep{cred2020}. Reliable flood
peak forecasting underpins early warning, infrastructure planning, and
emergency response \citep{rosenzweig2011}, yet operational deployment
remains constrained by trade-offs between model fidelity, computational
cost, and geographic transferability. Process-based hydrodynamic models,
from reach-scale codes such as HEC-RAS \citep{usace2021} and LISFLOOD-FP
(Bates and De Roo 2000) to the continental National Water Model
\citep{cosgrove2024}, resolve fluid dynamics explicitly but require
high-resolution terrain data, intensive parameter calibration, and
substantial computational resources, limiting applicability in
data-sparse basins and rapid-response contexts.

Data-driven alternatives have emerged as a promising complement, with
deep learning (DL) models \citep{bao2025,bentivoglio2025}
demonstrating competitive performance and substantially faster inference
than process-based benchmarks \citep{fraehr2023,shen2023}. A
growing body of evidence further suggests that data-driven networks can
outperform process-based models in generalization ability \citep{kratzert2024}. For example, an LSTM model maintained relatively high
accuracy when predicting extreme events compared with the US National
Water Model, a process-based benchmark\citep{frame2022}.

Remote sensing has become the principal source of observed flood
information at regional to global scales. Synthetic aperture radar (SAR)
and optical imagery are routinely used to delineate inundation extent
\citep{bonafilia2020,tellman2021,twele2016}, and
floodwater depth can be inferred by intersecting the mapped extent with
a DEM, as in the Floodwater Depth Estimation Tool (FwDET) \citep{cohen2018,cohen2019,peter2022}. These approaches describe floods that
have already occurred: revisit intervals and cloud cover mean that the
inundation peak is seldom imaged, and none of them can be issued before
the peak. Their depth estimates also carry a structural weakness. Depth
is obtained as the difference between an inferred water surface and the
DEM, so the vertical error of the DEM passes undiminished into the
depth: the global Copernicus GLO-30 DEM has a root-mean-square error of
about 4 m \citep{simard2024}, several times the depth that separates
nuisance from life-threatening flooding, and no extent-based method can
correct it because no observation of depth enters the computation. In
parallel, geospatial foundation models now compress multi-sensor EO
archives into compact, analysis-ready embeddings \citep{brown2025,cong2022,jakubik2023}, but whether such embeddings
carry information relevant to flood peak depth, and how they should be
combined with terrain and meteorological forcing, has not been tested.

Where depth is needed for response and recovery, agencies therefore
return to the field. In the United States, the flood depth grids that
FEMA and the USGS publish after major floods are not produced by
hydrodynamic models but by interpolating a water surface through the
high-water marks surveyed after the event and subtracting a lidar DEM
\citep{watson2018,musser2017}. These grids are operational
products: after Hurricane Sandy they identified 44,000 damaged and
inaccessible households and assigned flood depths to more than 200,000
flood-insurance policies \citep{bausch2015}, and FEMA guidance makes HWM
collection and depth-grid production a standard step of recovery (FEMA
2018). Their accuracy depends on the number and spacing of the marks
\citep{watson2018}, flooding away from the marks is missed \citep{tedesco2020}, and they become available only after the marks have been
surveyed, one to two weeks after landfall in the case of Sandy \citep{bausch2015}. A predictor that could place the water surface at such points
before the peak would make the same maps available ahead of the flood.

The quality of training data is a fundamental bottleneck for DL
performance. Currently, flood prediction datasets with high-quality
in-situ flood depth labels are scarce. Existing DL-based flood
prediction approaches \citep{fereshtehpour2024,liou2024,mirhoseini2024} predominantly rely on physics-based
hydrodynamic simulations as training supervision: these
simulation-trained surrogates inherit the systematic errors and spatial
biases of the underlying hydrodynamic models, and cannot capture the
local physical and anthropogenic processes that simulations fail to
represent. As a result, they struggle to generalize to real-world flood
conditions \citep{ursu2025} not reproduced by the simulator, and, to
our knowledge, no deep-learning flood depth model has yet been trained
and evaluated on field-observed peak depths: observations have served to
validate hydrodynamic models \citep{rogers2025} but not as supervision
for learning. Whether a network trained on observations alone can reach
useful accuracy, and how its errors behave at new events and in new
regions, is therefore unknown. Observational datasets such as river
gauges and high-water mark (HWM) surveys are spatially sparse \citep{blay2025,kratzert2019}, but provide accurate
ground-truth measurements of actual flood events. USGS HWM field surveys
record flood peak water surface elevation and inundation depth at
high-water mark locations, directly representing the real-world flood
conditions that simulation-based models approximate.

Three gaps follow. First, no benchmark pairs field-surveyed flood peak
depths with the time at which the peak was reached and with the Earth
observation (EO) and meteorological inputs available before it, so
observation-based learning of flood depth has not been attempted.
Second, it is not known whether a model supervised by surveyed depth can
compensate for the vertical error of the spaceborne DEM it is given,
which extent-based depth estimation cannot, nor whether static or annual
EO representations such as foundation-model embeddings add information
beyond terrain and forcing. Third, it has not been tested whether point
predictions issued before the peak can feed the interpolation procedure
that FEMA and the USGS apply to surveyed marks, and thus deliver
post-event-quality depth maps ahead of the flood.

Here we address these challenges by constructing a time-series
multimodal flood peak prediction dataset from 5689 USGS HWM observations
across 39 events in 24 U.S. states and Puerto Rico (2011--2025) paired
with 3185 site-matched rain-free samples and developing TCM-Former, a
TerraCross-Modal Fusion Transformer that predicts flood depth (height
above ground) as its primary target, with three-class flood hazard
severity and absolute peak elevation as auxiliary tasks. The framework
integrates GLO-30 digital elevation models (DEM), AlphaEarth satellite
embeddings , and a physically defined three-tier hydrometeorological
descriptor (58 physically defined features per 24-hour window) spanning
a 144-hour antecedent record. We evaluate the model at four lead times
(6, 12, 18 and 24 h), on random, event-disjoint and region-disjoint
splits, and attribute its skill to the three inputs with Shapley values.

The major contributions of this work are as follows:

1) We present, to our knowledge, the first flood depth benchmark and
deep-learning model built entirely on field observations: 5689 USGS HWM
surveys with sensor-derived peak times, site-matched rain-free windows,
GLO-30 and AlphaEarth patches and 144-h forcing. The benchmark also
allows simulation-trained surrogates to be tested against observed
depths for the first time.

2) We show that supervision by surveyed depth corrects the vertical
error of the spaceborne DEM implicitly: the model predicts depth at
surveyed marks with an MAE of 0.39 m although the only terrain it sees
is a surface model whose error is several metres, whereas
differencing water levels with the same DEM, as extent-based methods do,
yields an error of 0.93 m at the same marks even with the surveyed water
level. We quantify what each EO input contributes to this skill and how
the skill transfers to unseen events and regions.

3) We test whether point predictions issued 6 h before the peak can feed
the interpolation procedure that FEMA and the USGS apply to surveyed
marks. The resulting map of a held-out hurricane matches the post-event
map, but so does a map built from a constant water surface at the same
marks, which shows that depth grids built from sparsely surveyed points
are governed by the interpolation and the terrain rather than by the
accuracy of the points, and that what limits pre-peak depth mapping is
the transfer of depth skill to unseen storms rather than the mapping
procedure.

\hypertarget{study-area-and-data}{%
\section{Study area and data}\label{study-area-and-data}}

The benchmark is built on the USGS Short-Term Network Flood Event
Viewer, which archives the high-water marks and water-level sensor
records collected after major floods in the United States. HWM data
represent manually surveyed measurements of the highest flood trajectory
(i.e., the flood peak): some records capture only the elevation of the
flood trajectory, some record only the height above ground, and some
include both. In this study, only samples containing both measurements
were retained. However, HWM records include only the flag or found date
and the surveyed date, neither of which corresponds to the actual time
at which the flood peak occurred. For each flood event, we therefore
supplemented the HWM data with sensor-based records that provide the
precise timing of the flood peak. The primary sensor-based source is the
USGS Peak Summary dataset, a post-event summary of maximum observed
values derived from water level sensors, stream gauges, and
rapid-deployment gauges. For flood events without available peak summary
data, additional observed records were acquired, including water level
sensor data, wave height sensor data, and rapid-deployment gauge data,
all of which also provide the precise time of flood peak occurrence.

To assign each HWM record an approximate timestamp corresponding to its
flood peak, we searched for sensor-based records within a 10 km spatial
radius under the same flood event. The resulting dataset comprises
5689 HWM samples spanning 39 flood events from 2011 to 2025 and covering
24 states and Puerto Rico. To expose the model to non-flooded
conditions, each HWM site was additionally paired with a rain-free
sample: the same location and terrain inputs with a 144-h rain-free
meteorological window selected from AORC within ±8--60 days of the peak
(first \textless1 mm, then \textless5 mm of total precipitation; not
overlapping any other HWM peak at the site by ±7 days) and an assigned,
not surveyed, depth of 0 m (Level 1). The combined 8874 samples were
split at random by row into training (80\%, 7103), validation (10\%,
886) and test (10\%, 885) sets; an HWM and its rain-free sample may fall
in different splits by design, so that the model is evaluated on
same-site contrasts between rain and no rain. Fig. 1 illustrates the
spatial distribution of the dataset.

\noindent\makebox[\linewidth][c]{\includegraphics[width=6.5in,height=2.12778in]{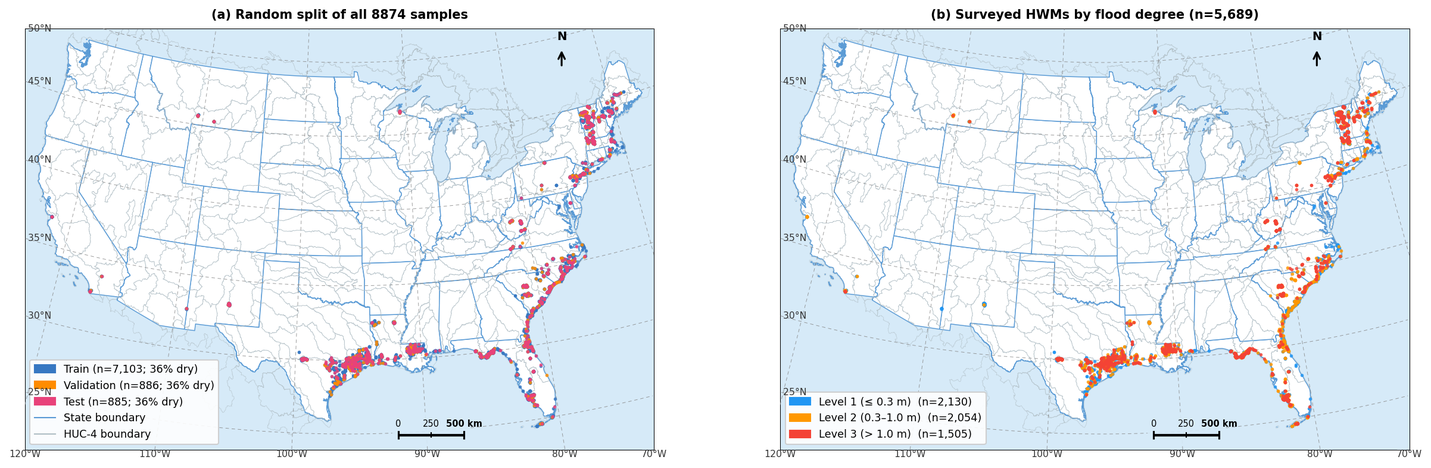}}

{\small
Fig. 1. Spatial distribution of the flooding dataset. (a) Spatial split
of dataset: Training 80\%, Validation 10\%, Testing 10\%, total:
8874 (5689 HWM + 3185 rain-free sample); (b) Flood degree distribution.
\par}

Flood peak elevation refers to the elevation of the flood peak
trajectory, while flood height above ground represents the flood depth,
that is, the difference between flood peak elevation and local ground
elevation. In this study, flood depth was classified into three
severity levels (0--0.3 m, 0.3--1.0 m, and \textgreater1.0 m) based on
hazard thresholds widely adopted in the literature. Depths below 0.3 m
are generally considered safe for vehicles, pedestrians, and buildings ,
and 0.3 m represents an established initial damage threshold
\citep{huizinga2017,sengphrachanh2021}. Depths
approaching 1.0 m mark the transition from low-impact nuisance flooding
to damaging and potentially life-threatening inundation, at which point
most furniture is submerged and electrical installations are heavily
damaged \citep{aryal2020,huizinga2017,sengphrachanh2021,smith2014}. Depths above 2.0 m correspond to an
extreme flood hazard threshold, beyond which flooding exerts severe
impacts on buildings, infrastructure, agriculture, and human safety
\citep{aryal2020,huizinga2017,komi2016,sengphrachanh2021,smith2014}. Depths above 2.0 m
were nevertheless merged into the highest class: they account for only
4.7\% of surveyed HWMs (269 samples) and could not be learned reliably
as a separate class in a four-class pilot, whereas the \textgreater1.0 m
class retains the operational meaning of vehicles floating and
ground-floor inundation. Based on the two remaining thresholds, flood
degree is divided into three levels according to flood depth, as
summarized in Table 1.

\begin{table}[!htbp]
\centering
\small
\caption{Flood severity levels used by the auxiliary classification task}\label{tab:1}
\begin{tabular}{@{}lccc@{}}
\toprule
Flood degree & Level 1 & Level 2 & Level 3 \\
\midrule
Depth $d$ & $d \leq 0.3$~m & $0.3~\text{m} < d \leq 1.0$~m & $d > 1.0$~m \\
\bottomrule
\end{tabular}
\end{table}

Fig. 2 shows the distribution of flood peak elevation, flood height
above ground, and flood degree. Flood peak elevation reflects
regional topographic variation, ranging from 0.2 m to about 2,300 m
(log\ensuremath{_{1}}\ensuremath{_{0}}(E) of −0.7 to 3.4), with 87\% of surveyed HWMs between 1 m and
316 m (log\ensuremath{_{1}}\ensuremath{_{0}}(E) of 0 to 2.5) and a long right tail reflecting site-level
topographic variation. Flood height above ground is more narrowly
distributed: among the 4220 HWMs with a non-zero depth, 97\% fall
between 0.1 and 10 m (log\ensuremath{_{1}}\ensuremath{_{0}}(H) of −1 to 1), with a median of 0.75 m and
a mode just below 1 m, indicating that most flood peaks reach moderate
inundation depths. Including the rain-free samples, Level 1 (depth
\ensuremath{\leq} 0.3 m) accounts for 59.9\% of samples, Level 2 for 23.1\% and Level 3
(depth \textgreater{} 1.0 m) for 17.0\%; among surveyed HWMs alone the
shares are 37.4\%, 36.1\% and 26.5\%, and 25.8\% of HWMs (1465) record a
depth of 0 m, i.e. the mark was found at ground level and the site is
treated as not inundated (Level 1).

\noindent\makebox[\linewidth][c]{\includegraphics[width=6.5in,height=1.77986in]{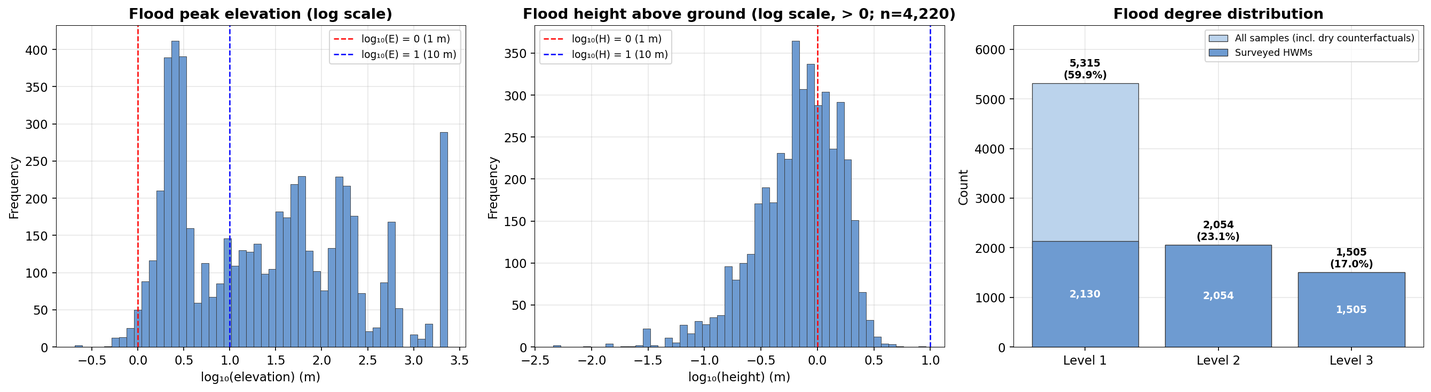}}

{\small
Fig. 2. The statistical distribution of flood peak elevation, flood
height above ground, and flood degree.
\par}

Flood conditions are governed by both static geographic factors and
dynamic hydrometeorological processes. To characterize the static
geographic context, we incorporated Digital Elevation Model (DEM) data.
To represent dynamic hydrometeorological forcing, we extracted 20
meteorological variables (as shown in Table A1 in Appendix A) covering
the period prior to the flood peak. Terrain captures the spatial
heterogeneity of flood impact within a given event, while meteorological
variables reflect the influence of varying atmospheric conditions on
flood depth at a specific location. To further enrich the spatial and
temporal context available for flood peak prediction, we
additionally incorporate site-level embeddings from the AlphaEarth (AE)
foundation model \citep{brown2025}as auxiliary spatial features.

AlphaEarth (AE) is a geospatial foundation model that fuses multi-modal
Earth observation data, including optical imagery, SAR remote sensing,
digital elevation, and climate signals, into compact, globally
consistent vector representations. Each embedding encodes rich surface
characteristics for a discrete geographic grid cell, capturing
information that extends beyond what individual spectral bands or
topographic variables can provide in isolation. These embeddings offer
several key advantages for flood peak prediction. First, they encode
static environmental and land-surface attributes, including terrain
morphology, land cover structure, and water body distribution, which are
fundamental determinants of potential inundation extent and maximum
flood depth. Second, because the embeddings are produced annually from
full-year image time series, they summarise the seasonal behaviour of
the surface, such as vegetation phenology and typical surface wetness,
which modulates runoff generation and inundation. By integrating AE
embeddings with time-series meteorological and topographical inputs, the
model benefits from complementary spatial context and dynamic
cross-modal linkages, improving the characterization of catchment
conditions and enhancing flood peak prediction under extreme rainfall
forcing.

For each sample, we extracted a GLO‑30 DEM tile centered on its
geographic coordinates, with a spatial extent of 255 × 255 pixels at a
30‑m resolution. Meteorological variables were obtained from the
Open‑Meteo API, from which we downloaded 20 core atmospheric
attributes; detailed descriptions are provided in the Appendix A.
Temporal context enters the model through the window position (each
of the six 24-h windows has its own linear embedding, and Tier 3
includes a normalized lookback position; Section 3.2) and through the
lead-time mask; the sample-level attributes time, latitude, longitude
and hours before the peak are not model inputs, and geographic
coordinates are used only to centre the spatial patches. Furthermore,
AE satellite embeddings with a spatial extent of 255 × 255 pixels at a
10 m resolution were also retrieved, centered on the same sample
coordinates.

Both spatial inputs are EO products. GLO-30 is the 30-m Copernicus DEM
derived from TanDEM-X X-band interferometric SAR acquired between 2011
and 2015; it is a digital surface model, so building and canopy heights
are retained, and its global RMSE is approximately 4 m \citep{simard2024}. AlphaEarth embeddings are 64-dimensional vectors at 10-m
resolution produced annually by the AlphaEarth Foundations model from
optical (Sentinel-2, Landsat), radar (Sentinel-1, ALOS PALSAR-2), lidar
(GEDI), elevation and climate inputs \citep{brown2025}. Because GLO-30
is itself one of the AlphaEarth inputs, the two modalities are partly
redundant; Section 4.2 quantifies the contribution of each. The DEM
patch (255 × 255 pixels, \ensuremath{\approx} 7.65 km) and the embedding patch (255 × 255
pixels, \ensuremath{\approx} 2.55 km) are centred on the same HWM but cover different
footprints, so the DEM branch sees catchment-scale relief while the
embedding branch focuses on the local land surface. AlphaEarth
embeddings are produced annually and are available from 2017 onward. No
sample uses the embedding of its own flood year: each HWM is paired with
the embedding of the year preceding its flood peak where one exists, and
otherwise with the first year after the event (2017 for the events of
2011--2016, 2018 for the 2017 events). Rain-free samples use the
embedding of their parent HWM. Test skill is not higher for samples with
post-event embeddings than for samples with pre-event embeddings
(macro-F1 0.613 versus 0.687 over all test samples, and 0.526 versus
0.635 over surveyed HWMs), so we find no indication that post-event
embeddings leak flood signatures, although the two groups also differ in
the events they contain. Table 2 summarizes all inputs.

\begin{table}[!htbp]
\centering
\small
\caption{Inputs of the benchmark, their sources and their role in TCM-Former.}\label{tab:2}
\begin{tabularx}{\textwidth}{@{}>{\raggedright\arraybackslash\hsize=0.77\hsize}X >{\raggedright\arraybackslash\hsize=1.33\hsize}X >{\raggedright\arraybackslash\hsize=0.92\hsize}X >{\raggedright\arraybackslash\hsize=0.92\hsize}X >{\raggedright\arraybackslash\hsize=1.06\hsize}X@{}}
\toprule
Input & Product and source & Spatial resolution and footprint & Temporal coverage & Role \\
\midrule
\multicolumn{5}{@{}l}{\textit{Labels and reference data}} \\
\addlinespace[2pt]
High-water marks & USGS Short-Term Network (STN) Flood Event Viewer surveys & Point & 39 events, 2011--2025 & Labels: peak elevation and depth; sample locations \\
\addlinespace
Peak timing & USGS peak summaries, water-level sensors and rapid-deployment gauges within 10 km of the HWM & Point & Per event & Assigns the flood peak time ($t = 0$) \\
\addlinespace
Ground elevation at the HWM & USGS 3DEP lidar-based elevation & Point & Static & Anchor of the residual elevation target (Section 3.3) \\
\addlinespace
Rain-free windows & NOAA Analysis of Record for Calibration (AORC) hourly precipitation & Gridded, nearest cell & $\pm$\,8--60 days around the peak & Selects the rain-free windows \\
\midrule
\multicolumn{5}{@{}l}{\textit{Model inputs}} \\
\addlinespace[2pt]
Terrain & Copernicus GLO-30 DEM (TanDEM-X, 2011--2015 acquisitions) & 30 m; $255 \times 255$ pixels ($\approx$\,7.65 km) & Static & DEM branch \\
\addlinespace
Land-surface embeddings & AlphaEarth Foundations, 64 bands (Google Earth Engine) & 10 m; $255 \times 255$ pixels ($\approx$\,2.55 km) & Annual; year preceding the event where available (see text) & AE branch \\
\addlinespace
Meteorological forcing & Open-Meteo historical weather API (reanalysis-based), 20 hourly variables (Table A1) & Reanalysis grid cell nearest the site ($\approx$\,9--25 km) & 144 h before the peak, masked to the lead time & Three-tier hydrometeorological descriptor (58 features per 24-h window) \\
\bottomrule
\end{tabularx}
\end{table}

\hypertarget{method}{%
\section{Method}\label{method}}

Accurate flood peak prediction is an inherently complex task, requiring
the integration of both static geographic context and dynamic
hydrometeorological forcing. As described above, the input data in our
dataset comprise DEM, AE embeddings, and time-series meteorological
data. To jointly exploit these complementary modalities, we developed
the TerraCross-Modal Fusion Transformer (TCM-Former), a multimodal
network whose primary output is flood depth and which predicts the flood
peak elevation and the flood degree as auxiliary tasks. The model
explicitly captures cross-modal interactions: DEM constrains
terrain-driven flow pathways, AE embeddings provide complementary
land-surface characteristics and spatial context at annual temporal
resolution, and time-series meteorological data represent the dynamic
hydrometeorological forcings that drive flood evolution.

As shown in Fig. 3, meteorological, DEM, and AlphaEarth (AE) inputs are
processed by three separate feature extractors. The AE features do not
explicitly represent local terrain. We therefore use cross-attention
to integrate terrain information from the GLO30 DEM. The AE tokens
serve as queries (Q), while the DEM tokens provide the keys (K) and
values (V). The DEM-enhanced AE embedding is then concatenated with the
DEM branch features and the meteorological features. The concatenated
features are fused by a multilayer perceptron (MLP) and passed to three
prediction heads. The primary head predicts the flood-peak height
above ground (flood depth); two auxiliary heads predict the flood-peak
elevation residual and three-class flood-peak degree logits. The final
flood-peak elevation is obtained by adding the predicted residual to the
local DEM elevation. The severity class is determined by the maximum
class logit.

\noindent\makebox[\linewidth][c]{\includegraphics[width=3.78704in,height=3.78704in]{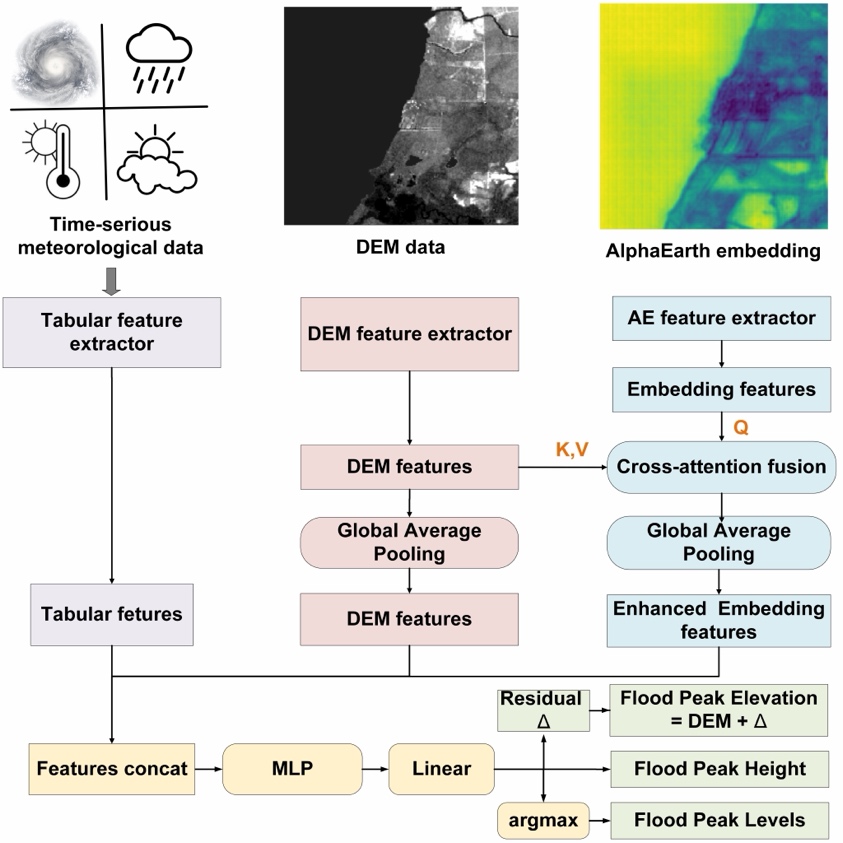}}

{\small\centering
Fig. 3. Multimodal flood prediction architecture.
\par}

\hypertarget{feature-extraction-and-processing-of-dem-and-ae-embeddings}{%
\subsection{Feature extraction and processing of DEM and AE embeddings}\label{feature-extraction-and-processing-of-dem-and-ae-embeddings}}

ViT-CoMer captures high-level, globally contextualized representations
through self-attention and convolution-enhanced multi-scale feature
interaction \citep{dosovitskiy2020}yielding semantically rich and
spatially coherent abstractions. In contrast, Convolutional Neural
Networks (CNNs), including DenseNet, emphasize local continuity,
fine-grained spatial detail, and feature reuse through densely connected
convolutional layers \citep{huang2017}, producing representations that
preserve high-resolution structural and texture-level information.

As illustrated in Fig. 4 and Fig. 5, the two spatial modalities, DEM and
AE, are processed by parallel CNN and ViT-CoMer \citep{xia2024}
branches. The CNN branch is used to extract local spatial features. For
DEM, DenseNet \citep{huang2017} is used to capture local topographic
patterns. For AE, a lightweight CNN is used to refine local spatial
patterns in the pretrained embeddings. In parallel, ViT-CoMer is used to
extract multi-scale contextual features. The features from the CNN and
ViT-CoMer branches are then fused through cross-attention.

\noindent\makebox[\linewidth][c]{\includegraphics[width=6.20833in,height=2.20833in]{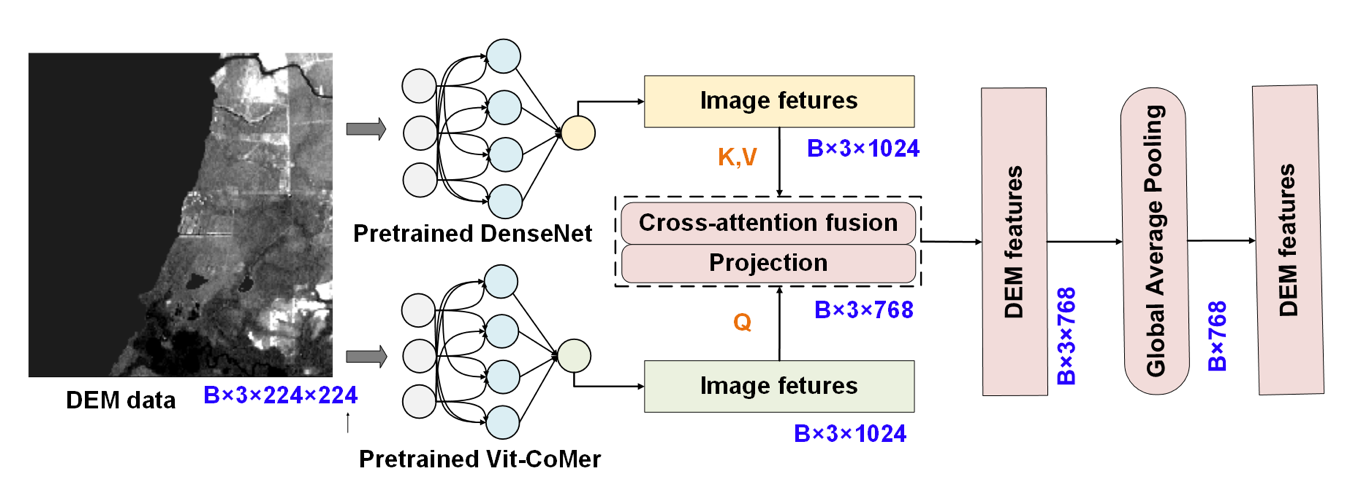}}

{\small\centering
Fig. 4. Feature extraction of DEM data
\par}

\noindent\makebox[\linewidth][c]{\includegraphics[width=6.5in,height=2.26597in]{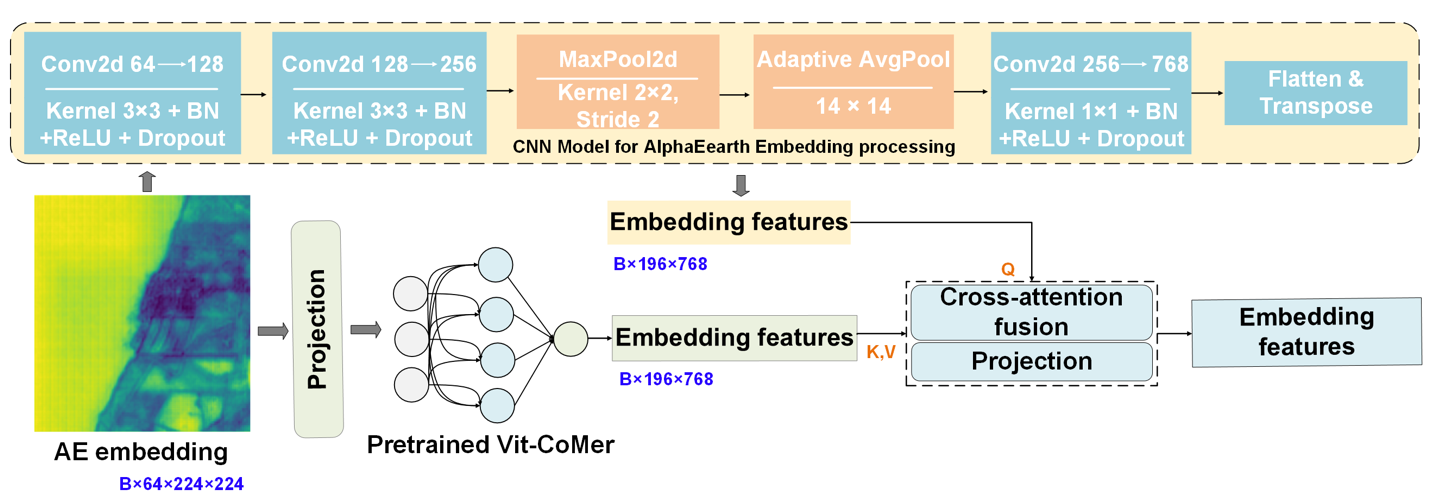}}

{\small\centering
Fig. 5. Feature extraction of AlphaEarth (AE) Embedding data
\par}

\hypertarget{feature-extraction-methods-for-time-series-meteorological-data}{%
\subsection{Feature extraction methods for time-series meteorological data}\label{feature-extraction-methods-for-time-series-meteorological-data}}

To extract informative representations from time-series meteorological
data, we adopt a hand-crafted three-tier hydrometeorological
descriptor (a set of physically defined features) to represent the
time-series meteorological data. It integrates physics-motivated domain
features (Tier 1), intra-window dynamics and coupling statistics (Tier
2), and cross-window antecedent condition descriptors (Tier 3), giving
58 features per 24-h window that characterize the forcing from
within-window structure to multi-day hydrometeorological evolution. The
features were compiled from established hydrometeorological indicators,
such as rainfall intensity statistics, reference evapotranspiration and
the antecedent precipitation index; each has an explicit physical
definition and is computed analytically without model inference, so that
the drivers of a prediction can be traced to hydrometeorological
processes. Two learned temporal encoders, described at the end of this
section, were evaluated as alternatives.

As shown in Fig. 6, for each HWM sample, the raw meteorological records
were first organized into a fixed 144-hour pre-peak sequence spanning
1--144 h before the assigned flood peak time. To simulate different
forecast lead times, we applied lead-time-specific exclusion before
feature extraction. Specifically, for an L-hour lead time, all
meteorological records within 1--L h before the flood peak were removed
before computing the three-tier meteorological features, and empty
near-peak windows were subsequently represented by zero-filled feature
vectors. This procedure ensures that Tier-1, Tier-2, and Tier-3 features
do not contain observed meteorological information from within the
forecast horizon, thereby preventing target leakage. In particular, the
Tier-3 cumulative features aggregate only antecedent windows outside the
forecast horizon.

\noindent\makebox[\linewidth][c]{\includegraphics[width=6.5in,height=4.96296in]{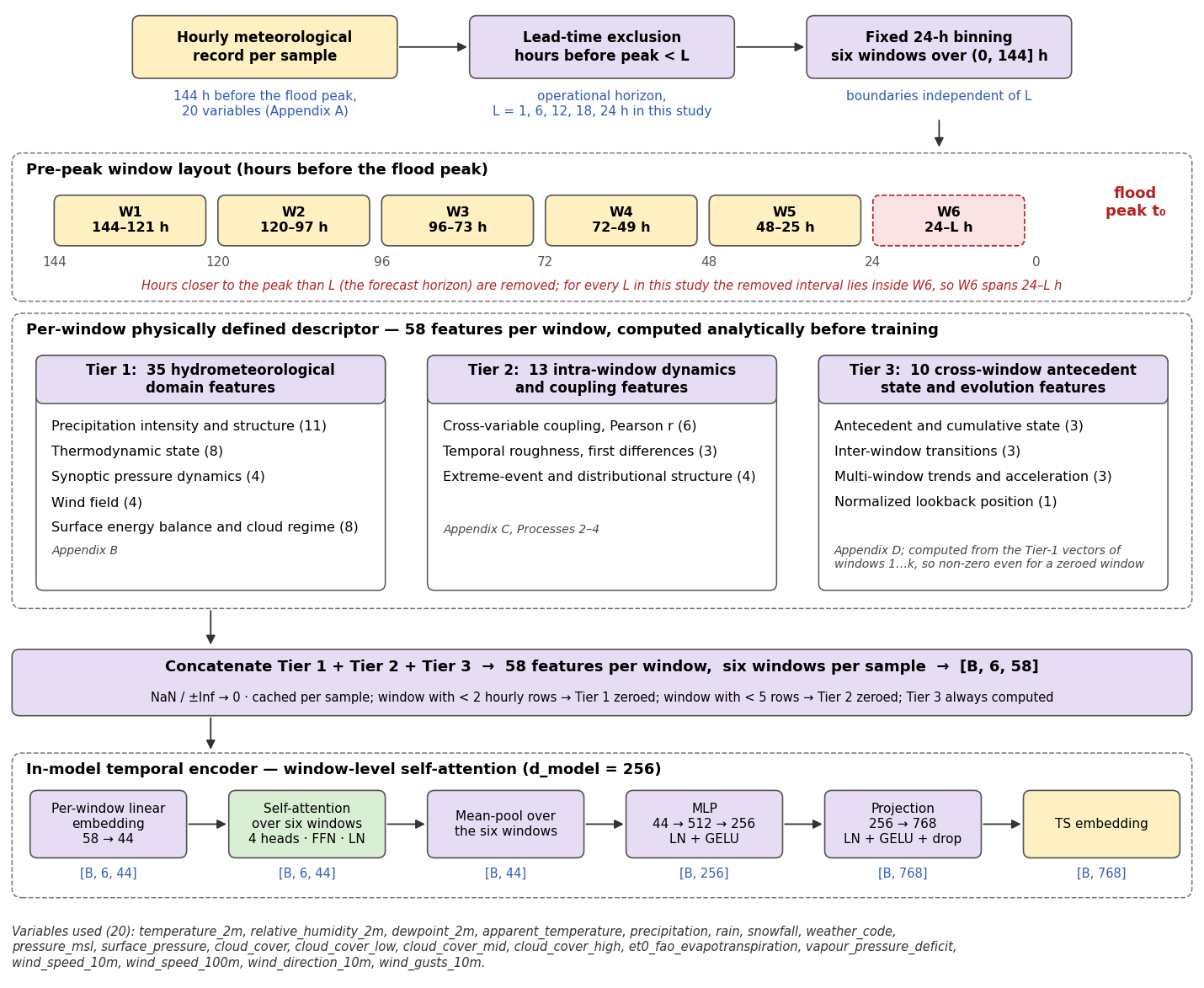}}

{\small
Fig. 6. Feature extraction from meteorological time-series data.
Meteorological records are divided into six 24-h windows and processed
using three tiers of feature extraction. The extracted features are
concatenated to form a 58-dimensional meteorological descriptor per
window. Invalid values (NaN or ±Inf) and features from windows with
insufficient observations are set to zero.
\par}

\textbf{Tier 1: Physics-based hydrometeorological domain features}

Tier 1 extracts 35 physically interpretable features directly from raw
meteorological observations within each 24-hour window, organized into
five thematic groups: (1) precipitation intensity and structure (11
features), characterizing the magnitude, variability, and temporal
organization of rainfall, including accumulated total, peak hourly rate,
wet-hour count, and peak 3-hour rolling accumulation; (2)
thermodynamic state (8 features), covering temperature, humidity,
and vapour pressure deficit, which govern evapotranspiration demand,
snowmelt potential, and near-surface atmospheric stability; (3)
synoptic pressure dynamics (4 features), describing mean sea-level
pressure and its trend, providing direct indicators of cyclonic
development and synoptic-scale forcing; (4) wind field (4 features),
capturing mean and maximum wind speed and their temporal trend, which
mediate moisture transport; and (5) surface energy balance and cloud
regime (8 features), including reference evapotranspiration ET\ensuremath{_{0}}, net
water input (P − ET\ensuremath{_{0}}), cloud cover, and WMO weather code, collectively
characterizing the surface moisture state and radiation environment
prior to flood onset.

These features are derived from the 20 meteorological variables in
Appendix A. All Tier 1 features are computed analytically without model
inference, as are the Tier 2 and Tier 3 features. Detailed
definitions of all 35 features are provided in Appendix B.

\textbf{Tier 2: Intra-window meteorological dynamics and coupling
features}

Tier 2 extracts 13 intra-window dynamic features from the
time-series meteorological data through three complementary processes
(Processes 2--4 in Appendix C; Process 1, based on the TabPFN-TS
forecaster, is not part of the descriptor and is evaluated separately as
a learned encoder). Process 2 encodes multivariate atmospheric coupling
via Pearson correlations between six physically motivated variable
pairs, characterizing whether the prevailing regime reflects organized
frontal or cyclonic forcing versus isolated convection. Process 3
quantifies temporal roughness via standard deviation of first
differences, distinguishes steady stratiform rainfall from intermittent
convective bursts. Process 4 characterizes extreme precipitation
structure through wet streak (precipitation \textgreater{} 0.1 mm h\ensuremath{^{-1}}),
Gini coefficient, and heavy-rain hour count (precipitation exceeding 10
mm h\ensuremath{^{-1}}), which are directly relevant to soil saturation and urban
drainage capacity thresholds. Full feature definitions and index mapping
are provided in Appendix C.

\textbf{Tier 3: Cross-window antecedent condition and temporal evolution
features}

Tier 3 encodes temporal memory and multi-day hydrometeorological
evolution through 10 cross-window features, aggregated across all
windows up to and including the current one. This tier is physically
motivated by the well-established role of antecedent soil moisture and
cumulative precipitation loading in flood generation. The 10 features
span four aspects: (1) antecedent cumulative state, including
accumulated precipitation, an Antecedent Precipitation Index (API, \ensuremath{\kappa} =
0.85) capturing soil moisture memory with exponential decay, and net
water surplus (P − ET\ensuremath{_{0}}); (2) inter-window transition dynamics, including
window-to-window differences in precipitation, MSLP trend, and relative
humidity; (3) multi-window secular trends, including least-squares
slopes of precipitation totals and MSLP; and (4) precipitation
acceleration and a normalized temporal position index, enabling the
model to weigh identical meteorological conditions differently depending
on their proximity to the flood event. Full mathematical definitions are
provided in Appendix D.

\textbf{Learned temporal encoders evaluated for comparison}

To test whether learned representations of the forcing carry
information that the three-tier descriptor misses, we replaced the
descriptor with two learned temporal encoders, keeping the spatial
branches, fusion module, prediction heads, training protocol and
lead-time masking unchanged. (1) TabPFN-TS: the pretrained TabPFN-TS
forecaster \citep{hoo2025}, which recasts forecasting as tabular
regression with TabPFN-v2, forecasts, in univariate mode, the next 6 h
of each of the 20 hourly variables (Table A1) from its series within
each 24-h window; the point forecasts 1, 3 and 6 h ahead and the 10\%
and 90\% quantiles 1 and 6 h ahead are retained, giving 140
model-derived features per window, standardized with training-set
statistics, in place of the 58 descriptors. (2) LSTM: a two-layer LSTM
\citep{hochreiter1997} with 128 hidden units reads the
standardized hourly series of all 20 variables over the 144-h record,
with precipitation, rain and snowfall log-transformed and a flag marking
the hours removed by the lead-time mask, and its final hidden state
replaces the descriptor.

\hypertarget{cross-modal-fusion-and-multi-task-heads}{%
\subsection{Cross-modal fusion and multi-task heads}\label{cross-modal-fusion-and-multi-task-heads}}

For cross-modality feature fusion, after modality-specific
CNN--ViT-CoMer encoding, the enhanced DEM and AE representations are
further fused through cross-attention. This final spatial fusion stage
integrates terrain-based flow constraints from DEM with land-surface
semantic context from AE embeddings. Given that AE aggregates
multi-source satellite image features over an annual temporal span, it
is necessary to incorporate DEM-based terrain information in a spatially
aligned manner, producing DEM-enhanced AE representations. DEM
features, enhanced AE embedding features, and tabular meteorological
features are then concatenated and fused through an MLP-based~fusion
mechanism. The fused representation is subsequently fed into
task-specific prediction heads: the flood height above ground (depth)
head, which gives the primary output, and two auxiliary heads for the
residual flood peak elevation and the three-class flood degree.

Because flood peak elevation is strongly correlated with local ground
elevation, we adopt a topographic prior residual learning strategy
(DEM-anchored elevation regression). Rather than predicting absolute
flood peak elevation directly, the model learns the residual between the
true flood peak elevation and the ground elevation at the HWM, thereby
exploiting the terrain prior while allowing the network to compensate
for terrain bias. The anchor is the USGS 3DEP lidar-based ground
elevation at the HWM location (Table 2), a point value that is more
accurate than the GLO-30 surface model; the GLO-30 patch enters the
model only through the DEM branch. The predicted flood peak elevation is
then recovered by adding the predicted residual to the same 3DEP ground
elevation. Residual labels are clipped to {[}−5, 10{]} m before the
variance-stabilizing transform so that a few survey outliers do not
dominate the fitted transform. The training objective comprises the
flood height above ground (depth) loss of the primary task (Lh) and two
auxiliary sub-losses, the flood peak elevation residual loss (L\ensuremath{_{r}}) and
the flood degree classification loss (L\ensuremath{^{c}}). The residual label is
computed according to Equation 1.

\(\Delta_{label} = {FE}_{label} - DEM\) (1)

where \(\Delta^{label}\) represents the flood peak elevation
residual, \({FE}^{label}\) represents the flood peak elevation label,
and \(DEM\) represents the 3DEP ground elevation at the HWM.

All models reported below use the residual formulation.

The overall training loss is a weighted combination of the three
sub-losses, with equal weights assigned to each task. All regression
targets are variance-stabilized and then standardized to zero mean and
unit variance using training-set statistics only, so that the
sub-losses are dimensionless and of comparable magnitude. And equal
weights express equal relative emphasis. Mean squared error (MSE) is
used as the loss function for both regression tasks (flood peak
elevation residual and flood height above ground). Cross-entropy loss
with inverse-frequency (balanced) class weights and label smoothing of
0.1 is used for flood degree classification. Depth accuracy, the primary
measure, is evaluated with the mean absolute error (MAE) and root mean
squared error (RMSE); the same metrics are used for the auxiliary
residual elevation, and micro-F1 (MIF1) and macro-F1 (MAF1) scores for
the auxiliary classification.

After training, the model achieving optimal performance on the
validation set is selected as the final model for subsequent testing.
The composite metric used for model selection across the three tasks is
defined by Equation 2. For the flood-degree classification task, the
MAF1 is adopted as the task metric: it is the unweighted mean of the
per-class F1 scores, so that a model is rewarded only if it classifies
the rare degrees as reliably as the common ones. Model selection is
therefore driven by balanced performance across all three degrees
rather than by the majority classes alone.

\(Metric = Lv + W_{Er} \times RMSE_{Er} + W_{H} \times RMSE_{H} - W_{D}\  \times MA{F1}_{D}\)
(2)

Where \(Lv\) is the validation loss, \(RMSE_{Er}\) is the RMSE for
residual based elevation, \(RMSE_{H}\) is the RMSE for flood height
above ground, and \(MAF_{D}\) is the macro f1 score for flood degree
classification, and \(W_{Er}\), \(W_{H}\), and \(W_{D}\) are the
corresponding task weights.

\hypertarget{evaluation-protocol}{%
\subsection{Evaluation protocol}\label{evaluation-protocol}}

All experiments were conducted on a single NVIDIA A100 or H100 GPU
(80 GB) with CUDA acceleration. The implementation was developed in
Python using PyTorch. Training ran for up to 100 epochs with early
stopping enabled; the batch size was 16 for both training and
evaluation. Gradient accumulation over 4 steps yielded an effective
batch size of 64. The base learning rate was set to 5 × 10\ensuremath{^{-5}}, with a
cosine-annealing schedule and 500 warm-up steps. A hierarchical learning
rate strategy was applied, in which the classifier learning rate was set
to 2 × 10\ensuremath{^{-4}}, four times the backbone learning rate. Optimization
employed AdamW with weight decay 0.1, \ensuremath{\varepsilon} = 1 × 10\ensuremath{^{-8}}, and a maximum
gradient norm of 1.0 for clipping. The model was evaluated on the
validation set every 25 optimizer steps; early stopping used a patience
of 15 epochs, and the optimal checkpoint was selected according to the
composite metric defined in Equation 2, with the five best checkpoints
retained. The three task weights in Equation 2 were set to 0.1, 0.1 and
3, respectively, to account for the differing magnitudes of the
individual indexes. All results are reported on the 885-sample test set
(567 surveyed HWMs and 318 rain-free samples) as the mean over three
random seeds (42, 1337 and 2024).

All metrics are reported as the mean over the three seeds;
seed-to-seed standard deviations are given in the table captions. For
the 6-h model, 95\% confidence intervals from 2000 bootstrap resamples
of the 885 test samples are 0.233--0.280 m for the depth MAE,
0.424--0.510 m for the depth RMSE, 0.702--0.753 for micro-F1 and
0.612--0.674 for macro-F1. The 318 rain-free samples have a depth of 0 m
and are predicted almost exactly (depth MAE 0.015 m), so the all-sample
depth error understates the error at flooded sites; where relevant we
therefore also report the depth error over the 567 surveyed HWMs.
Differences between configurations are tested with
Welch\textquotesingle s t-test over seeds where stated. To anchor the
metrics, Table 7 also lists two naive reference predictors evaluated on
the same test set: a majority-class classifier combined with a constant
depth (the training median, 0 m), and a class-prior random classifier
combined with the training mean depth (0.43 m).

\hypertarget{experimental-results}{%
\section{Experimental results}\label{experimental-results}}

\hypertarget{prediction-skill-across-forecast-lead-times}{%
\subsection{Prediction skill across forecast lead times}\label{prediction-skill-across-forecast-lead-times}}

In this section, we evaluate flood depth prediction, together with
the two auxiliary tasks, with the three-tier hydrometeorological
descriptor across four forecast lead times (6, 12, 18 and 24 hours).
Lead times shorter than 6 h are not considered. Flood peak times are
taken from the nearest water-level sensor within 10 km of the mark, and
flood peaks are not synchronous in space: a flow wave needs time to
propagate along a river \citep{allen2018}, and in small urban
catchments the local response to intense rainfall is far faster than
that of the river network \citep{yang2016}, so a mark may reach its
local maximum before or after the sensor that supplies its peak time.
The size of this offset is not quantified here, and the shortest lead
time we evaluate is therefore 6 h. All meteorological inputs end L
hours before the flood peak, so the prediction uses no information that
would be unavailable at that time. The peak time itself is taken as
given, as it would be in operation from a stage or discharge forecast.
The lead time is therefore the interval between the last observation
used and the peak, and the experiment measures the skill of forecasting
peak depth conditional on the timing of the peak.

Performance is assessed primarily on flood height above ground
(depth, H) with the MAE and RMSE, and secondarily on the two auxiliary
tasks, residual flood peak elevation (Er; MAE and RMSE) and flood degree
classification (micro-F1, MIF1; macro-F1, MAF1). Detailed results are
reported in Table 3 and Fig. 7.

\begin{table}[!htbp]
\centering
\small
\caption{Flood depth (primary task) and auxiliary-task skill of TCM-Former at different lead times with the three-tier hydrometeorological descriptor, and with the two learned temporal encoders at the 6-h lead time (mean over three seeds; s.d. \ensuremath{\leq} 0.04 m for errors and \ensuremath{\leq} 0.018 for F1 scores). H: flood depth (height above ground); Er: residual flood peak elevation; MIF1 and MAF1: micro- and macro-F1 of flood degree classification.}\label{tab:3}
\begin{tabular*}{\textwidth}{@{\extracolsep{\fill}}lccccccc@{}}
\toprule
\multirow{2}{*}{Extractor} & \multirow{2}{*}{Lead time} & \multicolumn{2}{c}{Depth $H$ (m)} & \multicolumn{2}{c}{Residual $E_r$ (m)} & \multicolumn{2}{c}{Severity level} \\
\cmidrule(lr){3-4}\cmidrule(lr){5-6}\cmidrule(lr){7-8}
& & MAE & RMSE & MAE & RMSE & MIF1 & MAF1 \\
\midrule
\multirow{4}{*}{Three-tier descriptor} & 6 h & 0.257 & 0.468 & 0.530 & 0.818 & 0.727 & 0.642 \\
 & 12 h & 0.280 & 0.486 & 0.528 & 0.805 & 0.718 & 0.643 \\
 & 18 h & 0.273 & 0.487 & 0.536 & 0.820 & 0.709 & 0.624 \\
 & 24 h & 0.271 & 0.489 & 0.543 & 0.816 & 0.717 & 0.639 \\
\midrule
TabPFN-TS & 6 h & 0.258 & 0.457 & 0.541 & 0.812 & 0.728 & 0.647 \\
LSTM encoder (LSTM+CNN-ViT) & 6 h & 0.267 & 0.461 & 0.545 & 0.814 & 0.722 & 0.653 \\
\bottomrule
\end{tabular*}
\end{table}

Fig. 7 summarizes the three tasks across lead times. Flood depth,
the primary output, is predicted with an MAE of 0.257 m and an RMSE of
0.468 m at the 6-h lead time. The depth error changes little with the
horizon: from 6 h to 24 h it increases by 0.014 m in MAE and 0.021 m in
RMSE, and none of the longer lead times differs significantly from 6
h (Welch\textquotesingle s t-test over three seeds, p = 0.08--0.18). We
therefore use the 6-h lead time, the shortest horizon that excludes the
near-peak sensor lag, as the reference configuration in the following
sections. Over the 567 surveyed HWMs alone, where the rain-free samples
no longer dilute the error, the depth MAE is 0.392 m at 6 h and
0.406--0.422 m at 12--24 h. The errors shrink towards the mean of the
training depths: at 6 h the model overestimates marks shallower than 0.3
m (bias +0.19 m for marks at ground level and +0.22 m for 0--0.3 m), is
nearly unbiased between 0.3 and 1.0 m (−0.05 m), and underestimates
deeper flooding (−0.51 m for 1--2 m and −1.48 m for the 27 marks deeper
than 2 m), so the largest absolute errors occur at the deepest sites.

The auxiliary tasks are similarly insensitive to the lead time. The
residual peak-elevation error stays within 0.528--0.543 m (MAE-Er),
because the residual between peak and ground elevation is governed
mainly by terrain rather than by the forcing, and severity
classification varies within the seed-to-seed spread (macro-F1
0.624--0.643, micro-F1 0.709--0.727).

\noindent\makebox[\linewidth][c]{\includegraphics[width=6.5in,height=1.90667in]{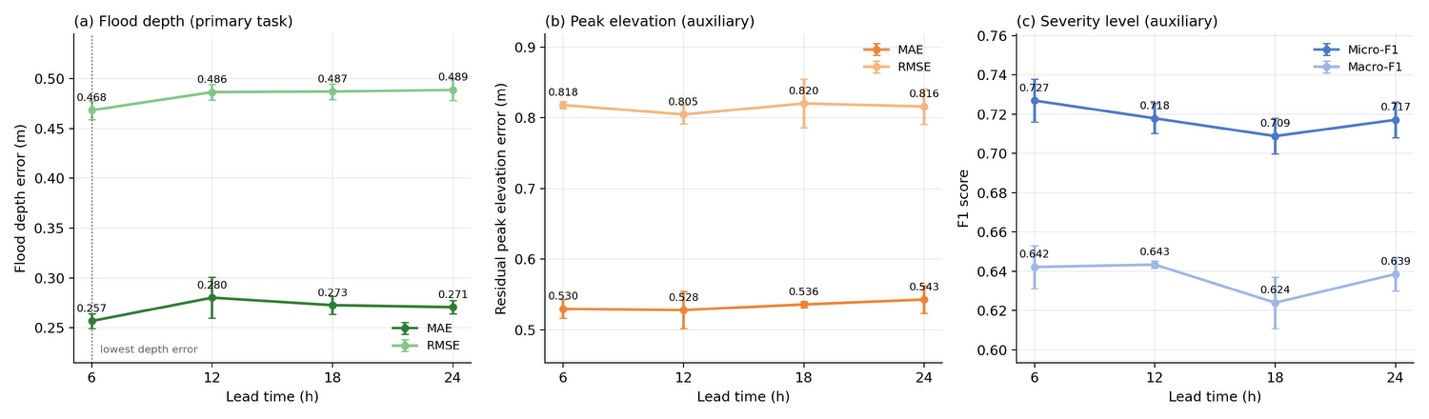}}

{\small
Fig. 7. Prediction skill across lead times (885 test samples; mean ±
s.d. over three seeds). (a) Flood depth (primary task), MAE and RMSE;
the 6-h lead time gives the lowest depth error. (b) Residual flood peak
elevation (auxiliary task), MAE and RMSE. (c) Flood degree
classification (auxiliary task), micro-F1 and macro-F1.
\par}

Table 3 also lists two learned temporal encoders, TabPFN-TS and an
LSTM, which replace the three-tier descriptor while the rest of the
model is unchanged (Section 3.2). Neither changes the depth error
significantly (depth MAE 0.258 and 0.267 m versus 0.257 m), and the
choice of the descriptor is discussed in Section 5.3.

\hypertarget{contribution-of-earth-observation-modalities}{%
\subsection{Contribution of Earth observation modalities}\label{contribution-of-earth-observation-modalities}}

Modality contributions are quantified with the three-tier
hydrometeorological descriptor at the 6-h lead time, the horizon with
the lowest depth error in Table 3, and averaged over three seeds.
Table 4 presents the depth error and the auxiliary-task skill for
all seven modality combinations.

\begin{table}[!htbp]
\centering
\small
\caption{Flood depth (primary task) and auxiliary-task skill for different modality combinations. MAE-H and RMSE-H: MAE and RMSE of flood depth (height above ground); MAE-Er and RMSE-Er: MAE and RMSE of the residual flood peak elevation; MIF1 and MAF1: micro- and macro-F1 of flood degree classification. 6-h lead time; mean over three seeds; errors in m.}\label{tab:4}
\begin{tabular*}{\textwidth}{@{\extracolsep{\fill}}lcccccc@{}}
\toprule
\multirow{2}{*}{Modalities} & \multicolumn{2}{c}{Depth $H$ (m)} & \multicolumn{2}{c}{Residual $E_r$ (m)} & \multicolumn{2}{c}{Severity level} \\
\cmidrule(lr){2-3}\cmidrule(lr){4-5}\cmidrule(lr){6-7}
 & MAE & RMSE & MAE & RMSE & MIF1 & MAF1 \\
\midrule
DEM, Meteo, AE & 0.257 & 0.468 & 0.530 & 0.818 & 0.727 & 0.642 \\
\midrule
DEM & 0.431 & 0.643 & 0.828 & 1.136 & 0.429 & 0.378 \\
AE & 0.410 & 0.625 & 0.801 & 1.174 & 0.474 & 0.426 \\
Meteo & 0.310 & 0.538 & 0.727 & 1.148 & 0.629 & 0.559 \\
DEM, Meteo & 0.269 & 0.485 & 0.550 & 0.848 & 0.724 & 0.644 \\
AE, Meteo & 0.270 & 0.483 & 0.677 & 0.997 & 0.721 & 0.654 \\
DEM, AE & 0.423 & 0.629 & 0.811 & 1.116 & 0.507 & 0.397 \\
\bottomrule
\end{tabular*}
\end{table}

Fig. 8 compares the depth error of the seven combinations,
separately for the surveyed HWMs and the rain-free samples. Meteorology
separates two groups. Every combination that contains it predicts the
rain-free samples almost exactly (depth MAE 0.012--0.022 m), whereas
combinations without it assign 0.20--0.25 m of water to the same sites
on dry days, because a flooded HWM and its rain-free sample share
identical terrain and embeddings and differ only in their forcing. Over
all 885 test samples, the full model reaches a depth MAE of 0.257 m.
Removing the meteorological forcing raises it to 0.423 m (+65\%;
Welch\textquotesingle s t-test p \textless{} 0.001), removing both the
DEM and the embeddings raises it to 0.310 m (+21\%; p = 0.033), and
removing only the DEM or only the embeddings raises it to 0.270 or 0.269
m (+5\%; p = 0.35 and 0.10). Expressed as Shapley values, the three
inputs reduce the depth MAE from 0.416 m for a constant zero-depth
prediction to 0.257 m, of which meteorology contributes 0.141 m (88\%),
the embeddings 0.014 m (9\%) and the DEM 0.004 m (3\%). The spatial
inputs act where flooding occurs: they leave the rain-free samples
unchanged, whereas at the 567 surveyed HWMs (including 149 marks with a
surveyed depth of zero) removing both raises the depth MAE from 0.392 to
0.478 m (+22\%; p = 0.025); conversely, adding terrain and embeddings to
the forcing lowers the depth error at the surveyed marks by 18\%. The
DEM and the AE embeddings are interchangeable for depth: DEM +
Meteorological and AE + Meteorological reach HWM depth MAEs of 0.407 and
0.412 m, neither differs significantly from the full model (p = 0.07 and
0.37), and combining both gives a further, non-significant reduction.
Without meteorology, the spatial inputs, singly or together, give HWM
depth MAEs of 0.525--0.535 m. The auxiliary tasks follow the same
pattern for severity (macro-F1 0.64--0.65 with meteorology, below 0.43
without), but terrain is indispensable for the peak elevation: removing
the DEM raises MAE-Er from 0.530 to 0.677 m. In short, the
meteorological forcing carries most of the depth skill, including the
separation of flooded from rain-free conditions that the paired design
makes trivially learnable, and the spatial inputs reduce the depth
error at the surveyed marks by about a fifth.

\noindent\makebox[\linewidth][c]{\includegraphics[width=6.5in,height=2.42667in]{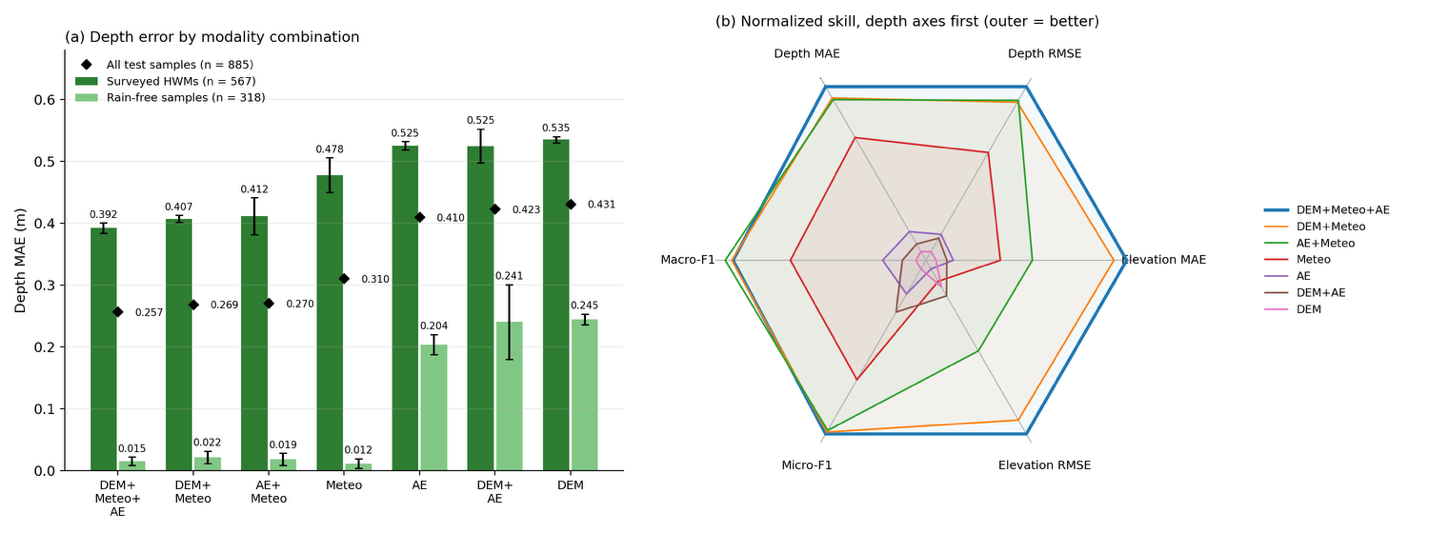}}

{\small
Fig. 8. Flood depth error and overall skill of the seven modality
combinations (6-h lead time; mean ± s.d. over three seeds). (a) Depth
MAE on the 567 surveyed HWMs and the 318 rain-free samples (bars) and on
all 885 test samples (diamonds). (b) Normalized scores of the six
metrics, depth axes first; outer values are better.
\par}

To better understand the contribution of each modality, we
calculated Shapley values \citep{lundberg2017,shapley1953} for the
depth MAE (MAE-H) and, for the auxiliary tasks, MAE-Er and MAF1, as
shown in Fig. 9. The Shapley value of each modality represents its
average marginal contribution across all possible modality subsets. For
classification, the payoff was defined as MAF1, with a majority-class
predictor as the no-modality baseline. For regression, the payoff was
defined as the reduction in MAE relative to a no-modality baseline that
predicts a constant (the training median depth and the training mean
residual), so that positive Shapley values consistently indicate
performance improvement. As shown in Fig. 9a, meteorology has the
largest Shapley value for depth (a depth MAE reduction of 0.141 m over
all samples), as it has for the auxiliary tasks (0.192 m for MAE-Er and
0.267 for MAF1), reflecting its role in separating flooded from
rain-free conditions. The Shapley values of the spatial inputs for depth
are small (0.004 m for the DEM and 0.014 m for the embeddings) because
the average runs over subsets without meteorology, in which the
rain-free samples dominate the depth error; conditional on meteorology,
adding the spatial inputs reduces the depth MAE at the surveyed HWMs by
0.086 m (Fig. 8a), which is where their contribution to flood depth
lies. For the auxiliary tasks the spatial inputs matter more: AE
embeddings contribute more to flood degree classification (0.077 versus
0.048 for the DEM), while the DEM contributes more to the residual flood
peak elevation (0.075 m versus 0.025 m).

Restricting the evaluation to the 567 surveyed HWMs (Fig. 9b) leaves
meteorology dominant for depth (0.098 m, versus 0.013 m for the DEM and
0.016 m for the embeddings): even among flooded sites, the amount of
water delivered by the storm sets most of the depth, and the spatial
inputs adjust it locally. For the auxiliary severity task the ranking
narrows on these sites (macro-F1 Shapley values of 0.158 for
meteorology, 0.126 for the embeddings and 0.106 for the DEM), and
meteorology alone is the weakest single input (macro-F1 0.399, versus
0.449 for the DEM and 0.494 for the embeddings), so the EO inputs carry
most of the information that separates severity levels among flooded
sites.

\noindent\makebox[\linewidth][c]{\includegraphics[width=6.5in,height=2.22857in]{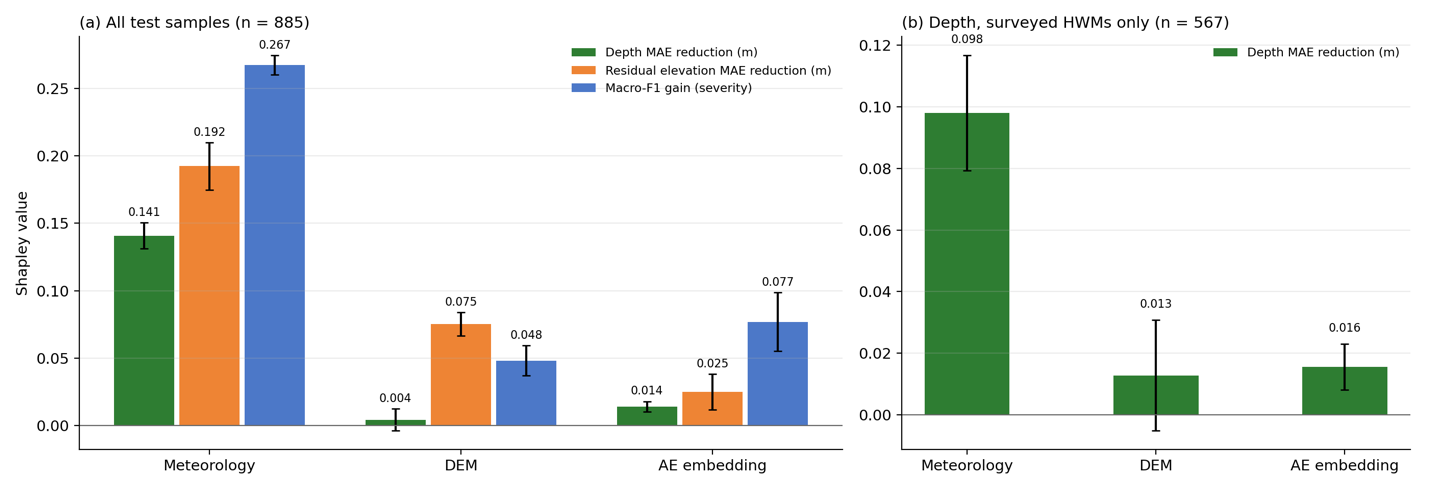}}

{\small
Fig. 9. Modality Shapley values (6-h lead time; error bars:
s.d. over three seeds). (a) All 885 test samples: depth MAE reduction
(primary task), and residual elevation MAE reduction and macro-F1 gain
(auxiliary tasks), relative to constant or majority-class predictors.
(b) Depth MAE reduction over the 567 surveyed HWMs, relative to a
constant prediction of the training median HWM depth (0.51 m).
\par}

\hypertarget{transferability-to-unseen-events-and-regions-and-baseline-comparison}{%
\subsection{Transferability to unseen events and regions, and baseline comparison}\label{transferability-to-unseen-events-and-regions-and-baseline-comparison}}

To test the ability of the model to generalize, we evaluated it on
unseen flood events and on unseen geographic sub-regions, using the same
8874 samples (5689 HWMs from 39 flood events and 3185 rain-free
samples), the three-tier hydrometeorological descriptor and the 6-h lead
time. Each rain-free sample is assigned to the same partition as its
parent HWM. For the event split, whole events were assigned to the
validation and test sets until each held about 10\% of the samples;
events contributing more than 5\% of all samples (e.g., 2017 Harvey)
were kept in training, and among 5,000 random event orderings we
selected the one whose validation and test sets best matched the 10\%
target and the overall share of dry samples (35.9\%). The resulting test
set contains 897 samples (575 HWMs) from 11 events, including 2018
Florence, 2017 Maria and 2024 Helene, and the validation set contains
892 samples from 13 events (Fig. 10).

\noindent\makebox[\linewidth][c]{\includegraphics[width=6.5in,height=2.56944in]{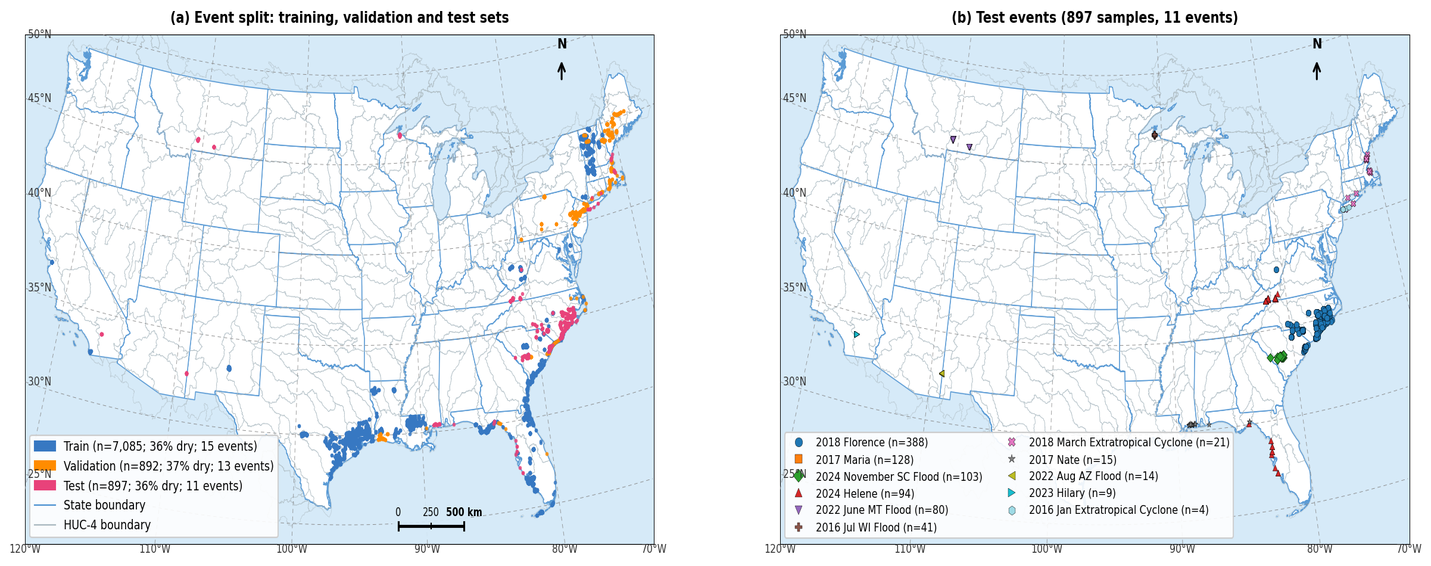}}

{\small
Fig. 10. Dataset splitting with flood events isolated for each
dataset. (a) The distribution of training, validation, and testing
dataset; (b) The distribution of different flood events on testing
dataset. Rain-free samples share the coordinates of their parent
HWMs; the 128 test samples of 2017 Maria (Puerto Rico) lie outside the
map extent.
\par}

Table 5 compares the event split with the random split. Skill
decreases for unseen events: the depth MAE rises from 0.257 to 0.386 m
(RMSE from 0.468 to 0.648 m), and over the surveyed HWMs from 0.392 to
0.574 m, with depths underestimated on average (bias −0.32 m). The
auxiliary tasks degrade similarly: the residual elevation MAE rises from
0.530 to 0.997 m, and macro-F1 falls from 0.642 to 0.537 (±0.012 over
seeds). Fig. 10 shows that many validation and test samples are
spatially separated from the training samples and, in some cases,
located in different HUC-4 regions, so the event split tests new storms
at largely new locations. The loss of skill is not confined to the
magnitude of the error. Over all 897 test samples the model retains
positive skill relative to a constant prediction (skill score 1 −
MSE(model)/MSE(constant) of +0.14 against the training mean and +0.43
against the training median), because it still separates flooded from
rain-free conditions, although the paired design makes that separation
largely learnable from the forcing alone. Among the 575 surveyed marks
it does not: the predicted depths occupy a narrow band, with a
regression slope of 0.06 against the surveyed depth, a coefficient of
determination of 0.03, a within-event correlation of 0.11 and a negative
skill score (−0.20 against the training mean, −0.07 against the training
median). On the random split the same quantities are a slope of 0.36, a
coefficient of determination of 0.29, a within-event correlation of 0.41
and skill scores of +0.23 and +0.26. Neither the magnitude nor the
ranking of depth therefore transfers to new storms, whereas within
sampled events and regions both are learned. Event-level transfer thus
remains far more challenging than random sample-level generalization, in
which HWMs from the same event and neighbourhood occur in both training
and test sets.

\begin{table}[!htbp]
\centering
\small
\caption{Generalization of TCM-Former to unseen flood events (6-h lead time; mean over three seeds; errors in m). The random split is the test set of Table 3 (885 samples); the event split holds out 11 whole events (897 samples).}\label{tab:5}
\begin{tabular*}{\textwidth}{@{\extracolsep{\fill}}lcccccc@{}}
\toprule
\multirow{2}{*}{Dataset} & \multicolumn{2}{c}{Depth $H$ (m)} & \multicolumn{2}{c}{Residual $E_r$ (m)} & \multicolumn{2}{c}{Severity level} \\
\cmidrule(lr){2-3}\cmidrule(lr){4-5}\cmidrule(lr){6-7}
 & MAE & RMSE & MAE & RMSE & MIF1 & MAF1 \\
\midrule
Random split & 0.257 & 0.468 & 0.530 & 0.818 & 0.727 & 0.642 \\
Event split & 0.386 & 0.648 & 0.997 & 1.489 & 0.614 & 0.537 \\
\bottomrule
\end{tabular*}
\end{table}

To investigate model transferability to unseen geographic sub-regions at
varying spatial separation distances from the training data, we divide
each HUC‑4 watershed into spatial cells of increasing fineness.
Specifically, we evaluate four variants, 4-intra (2×2 grid), 8-intra
(4×2), 16-intra (4×4), and 32-intra (8×4), each producing an
increasingly fine-grained spatial partition of the same 8874 samples.
In all variants, the grid boundaries are defined by quantiles of the
observed sample coordinates within each watershed, so that grid cells
adapt to the actual spatial distribution of measurement points rather
than fixed geographic coordinates.

Within each spatial cell, a greedy algorithm is applied: cells are
iteratively added to the validation set starting from the north-eastern
corner of each watershed, and to the test set starting from the
south-western corner, until each accumulates approximately 10\% of the
watershed\textquotesingle s samples. All remaining cells are assigned to
the training set. Cells containing fewer than three samples are
unconditionally assigned to training to prevent degenerate splits.
Rain-free samples share the coordinates of their parent HWMs and
therefore fall in the same cell. The resulting test sets contain 507
(4-intra), 1127 (8-intra), 1197 (16-intra) and 1077 (32-intra) samples.
Fig. 11 shows the spatial distribution of different splitting
scenarios. This procedure guarantees that training, validation, and
testing samples occupy spatially disjoint sub-regions within every HUC‑4
unit, with no geographic overlap across splits.

\noindent\makebox[\linewidth][c]{%
\includegraphics[width=0.49\linewidth]{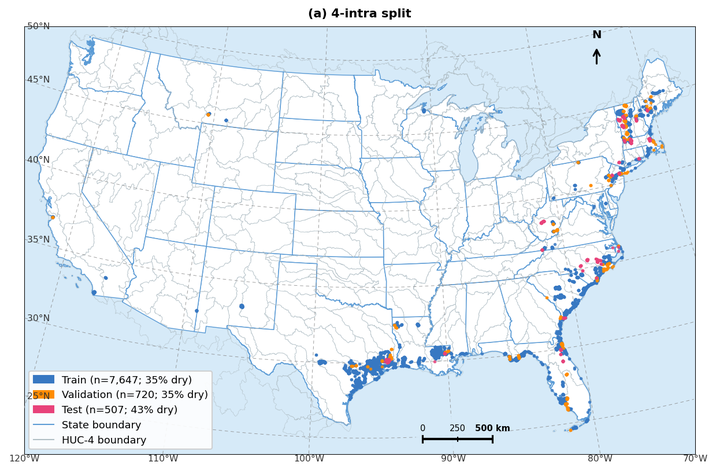}\hfill\includegraphics[width=0.49\linewidth]{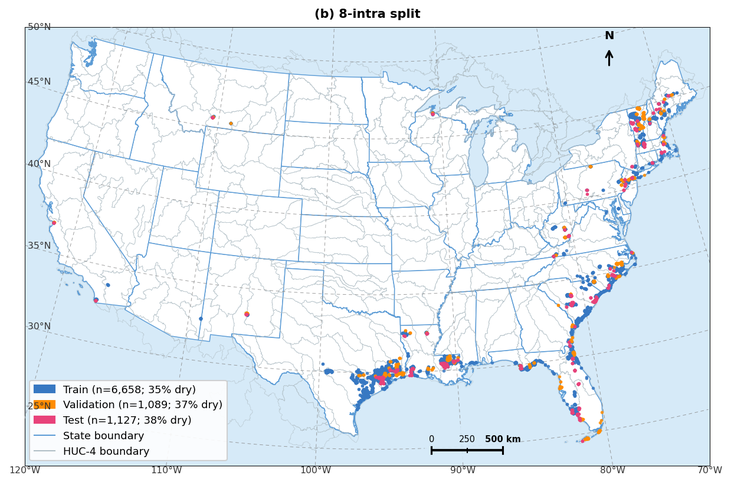}}

\noindent\makebox[\linewidth][c]{%
\includegraphics[width=0.49\linewidth]{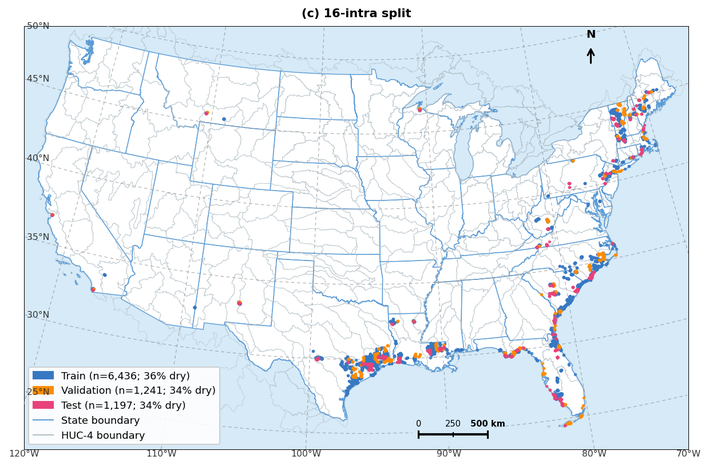}\hfill\includegraphics[width=0.49\linewidth]{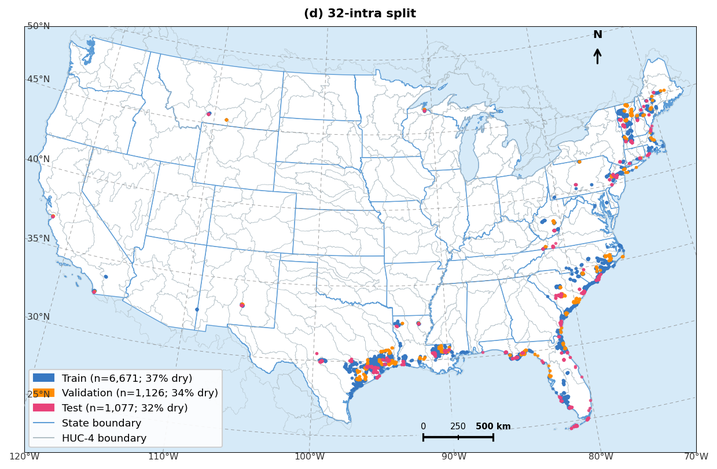}}

{\small
Fig. 11. Dataset splitting with different spatial regions isolated
for each dataset. (a) The distribution of training, validation, and
testing dataset for 4-intra splitting scenario; (b) for the 8-intra
splitting scenario, (c) for the 16-intra splitting scenario; (d) for the
32-intra splitting scenario. Rain-free samples share the coordinates
of their parent HWMs; legend shares are the fraction of dry samples in
each set.
\par}

Table 6 summarizes transferability to spatially disjoint
sub-regions. All four partitions reduce skill relative to the random
split, to a degree similar to the event split: the depth MAE ranges from
0.356 to 0.374 m (0.497--0.629 m over the surveyed HWMs), and macro-F1
from 0.550 to 0.577. Skill does not improve as the partition becomes
finer, although finer partitions place test cells closer to training
cells: the depth MAE varies without a trend (0.374, 0.359, 0.361 and
0.356 m from 4-intra to 32-intra), as does macro-F1, and micro-F1
decreases slightly from 0.655 to 0.623. The differences between
partitions (at most 0.018 m in depth MAE and 0.027 in macro-F1) are
comparable to the seed-to-seed standard deviation (up to 0.038 in
macro-F1), so they more likely reflect which sub-regions fall into each
test set than a systematic effect of distance. Residual elevation errors
(MAE-Er = 0.775--0.982 m) remain well above those of the random split
(0.530 m), consistent with the DEM precision ceiling discussed in
Section 5.2: a plausible explanation is that GLO-30 elevation errors are
spatially correlated and can be partly learned from neighbouring HWMs in
a random split, but not when whole sub-regions are held out. Together
with the event split, these results indicate that the model interpolates
within sampled events and regions better than it extrapolates to new
ones.

\begin{table}[!htbp]
\centering
\small
\caption{Generalization of TCM-Former to spatially disjoint sub-regions within HUC-4 watersheds (6-h lead time; mean over three seeds; errors in m).}\label{tab:6}
\begin{tabular*}{\textwidth}{@{\extracolsep{\fill}}lcccccc@{}}
\toprule
\multirow{2}{*}{Sampling} & \multicolumn{2}{c}{Depth $H$ (m)} & \multicolumn{2}{c}{Residual $E_r$ (m)} & \multicolumn{2}{c}{Severity level} \\
\cmidrule(lr){2-3}\cmidrule(lr){4-5}\cmidrule(lr){6-7}
 & MAE & RMSE & MAE & RMSE & MIF1 & MAF1 \\
\midrule
4-intra & 0.374 & 0.649 & 0.982 & 1.423 & 0.655 & 0.550 \\
8-intra & 0.359 & 0.582 & 0.775 & 1.090 & 0.650 & 0.577 \\
16-intra & 0.361 & 0.628 & 0.903 & 1.273 & 0.646 & 0.557 \\
32-intra & 0.356 & 0.620 & 0.787 & 1.128 & 0.623 & 0.562 \\
\bottomrule
\end{tabular*}
\end{table}

\begin{table}[!htbp]
\centering
\small
\caption{Comparison of TCM-Former and baseline models at the 6-h lead time (random split; mean over three seeds; TFT depth predictions clipped at zero, as in its loss).}\label{tab:7}
\begin{tabular*}{\textwidth}{@{\extracolsep{\fill}}>{\raggedright\arraybackslash}p{4.2cm}ccccccc@{}}
\toprule
\multirow{2}{*}{Methods} & \multirow{2}{*}{Lead time} & \multicolumn{2}{c}{Depth $H$ (m)} & \multicolumn{2}{c}{Residual $E_r$ (m)} & \multicolumn{2}{c}{Severity level} \\
\cmidrule(lr){3-4}\cmidrule(lr){5-6}\cmidrule(lr){7-8}
& & MAE & RMSE & MAE & RMSE & MIF1 & MAF1 \\
\midrule
Ours (TCM-Former) & 6 h & 0.257 & 0.468 & 0.530 & 0.818 & 0.727 & 0.642 \\
TFT & 6 h & 0.261 & 0.456 & 0.759 & 1.152 & 0.694 & 0.632 \\
LSTM+CNN-ViT & 6 h & 0.267 & 0.461 & 0.545 & 0.814 & 0.722 & 0.653 \\
\midrule
Majority class + constant depth (training median) & --- & 0.416 & 0.744 & --- & --- & 0.599 & 0.250 \\
\addlinespace
Class-prior random classifier + training mean depth & --- & 0.487 & 0.617 & --- & --- & 0.440 & 0.332 \\
\bottomrule
\end{tabular*}
\end{table}

We benchmark TCM-Former against two baselines at the 6-h lead time.
The first is a Temporal Fusion Transformer (TFT) \citep{lim2021},
which jointly processes static catchment properties and dynamic
meteorological sequences. 146 static indicators (latitude, longitude,
ground elevation, and statistics extracted from DEM and AE embeddings
such as slope and elevation statistics) are provided as static
covariates, while the hourly meteorological time series is processed by
an LSTM encoder within the TFT framework; its elevation target is the
same standardized residual used by TCM-Former. TCM-Former has a slightly
lower depth MAE than TFT, 0.257 versus 0.261 m over all samples and
0.392 versus 0.407 m over the surveyed HWMs (2\% and 4\% lower),
although neither difference is significant (Welch\textquotesingle s
t-test over three seeds, p = 0.51 and 0.16) and the depth RMSE of TFT is
lower (0.456 versus 0.468 m; the TFT loss clips negative depths at zero
and its predictions are evaluated accordingly). TFT has a 43\% larger
residual elevation error (MAE-Er 0.759 versus 0.530 m), a comparable
macro-F1 (0.632 versus 0.642) and a lower micro-F1 (0.694 versus
0.727), indicating that flat tabular covariates support depth
estimation at the surveyed sites but not the terrain-dependent peak
elevation. For the primary target, the advantage of the proposed
architecture over a strong tabular model given the same EO information
is therefore small; its clear advantage lies in the terrain-dependent
peak elevation, and the depth skill reported in this paper does not
hinge on the choice of network.

To isolate the contribution of our spatial feature extraction module, we
construct an ablation baseline (LSTM+CNN-ViT) that retains LSTM-based
temporal modeling while replacing the flat tabular static inputs with
our CNN--ViT cross-attention spatial representations. This
substitution raises micro-F1 from 0.694 to 0.722 and reduces MAE-Er from
0.759 to 0.545 m, whereas the depth MAE is unchanged (0.261 versus 0.267
m): the cross-modal spatial encoding mainly benefits the
terrain-dependent peak elevation.

Because LSTM+CNN-ViT shares the cross-modal spatial encoder of
TCM-Former and differs from it only in the meteorological branch, it is
the LSTM variant of TCM-Former evaluated in Section 4.1 (Table 3). Its
skill is statistically indistinguishable from that of TCM-Former (Table
7; Welch\textquotesingle s t-test over three seeds, p \textgreater{} 0.1
for all six metrics), showing that the gains over TFT come from the
cross-modal spatial encoding rather than from the meteorological
encoder.

The naive predictors in Table 7 put these numbers in context: the
two naive predictors reach depth MAEs of 0.416 m (constant training
median, 0 m) and 0.487 m (training mean, 0.43 m) and macro-F1 scores of
0.250 and 0.332, against 0.257 m and 0.642 for TCM-Former.

\hypertarget{event-scale-depth-maps-from-point-predictions}{%
\subsection{Event-scale depth maps from point predictions}\label{event-scale-depth-maps-from-point-predictions}}

The post-event depth grids described in the Introduction are produced by
interpolating a water surface through the surveyed marks and subtracting
a lidar DEM \citep{watson2018}. We ask whether the water surface
predicted before the peak, passed through the same procedure, yields a
map equivalent to the one built afterwards from the surveyed marks.

We applied the procedure to our point predictions for the held-out
event, 2018 Florence, on the Lumber River at Lumberton, where the river
left its floodplain and flooded the town: 28 HWMs of the event-split
test set lie in the reach, 15 of them in built-up areas (impervious
cover of at least 20\% within 500 m), 0.7--2 km from the channel. Map A
uses the surveyed peak elevation at each HWM; map B uses the predicted
water surface, i.e. the 3DEP ground elevation at the mark plus the depth
predicted 6 h before the peak by the event-split model (mean of three
seeds). Both point sets are interpolated with a linear TIN inside the
convex hull of the marks, the 3DEP 10 m bare-earth DEM (NAVD88) is
subtracted, and cells with positive depth are kept when connected to a
component that contains a mark. A and B therefore differ only in whether
the water surface at the marks is surveyed or predicted; because the
marks are known only after the event, this is a controlled comparison
rather than a forecast product. As an external reference we use the NC
OneMap Hurricane Florence flood extent \citep{schaffersmith2020}, a
20 m binary map of inland flooding obtained by random-forest
classification of pre- and post-storm Sentinel-1 radar imagery with
topographic, floodplain and land-cover predictors; it was trained on
USGS and NCDEM high-water marks and on NOAA aerial photographs and
reaches 92\% accuracy on a withheld sample according to its
documentation. The reference is therefore not independent of the marks,
carries no depth, and inherits the weakness of radar water detection
under tree canopy and between buildings, so we treat it as a reference
rather than as ground truth.

Fig. 12 shows the result. The predicted map reproduces the surveyed
one closely (Fig. 12a, b; extent CSI 0.85; 53 versus 49 km\ensuremath{^{2}} inundated
inside the 84 km\ensuremath{^{2}} hull; mean depth 1.20 versus 1.23 m; depth MAE 0.34 m
and bias +0.05 m where both are inundated). Against the reference, which
marks 40.4 km\ensuremath{^{2}} of the hull as flooded, the two maps are equivalent (Fig.
12c): 83\% of the reference flooded area is inundated in both maps, 7\%
in the predicted map only (2.9 km\ensuremath{^{2}}), 3\% in the surveyed map only (1.2
km\ensuremath{^{2}}) and 7\% in neither (2.7 km\ensuremath{^{2}}), giving a recall of 0.90 for B and
0.86 for A at a precision of 0.69 and 0.70 (CSI 0.64 and 0.63). Of the
16.6 km\ensuremath{^{2}} that B inundates but the reference does not, 13.7 km\ensuremath{^{2}} are also
inundated in A; this area lies mostly in the built-up centre of
Lumberton between the marks, where the reference is flooded within 60 m
of 24 of the 27 wet marks but shows little water between them, so the
disagreement is between a TIN that fills every low cell and a radar
product that detects water poorly among buildings, and the present data
cannot decide which is closer to the truth. The predicted depth in this
area is moderate (median 0.55 m, mean 0.61 m, s.d. 0.43 m): 24\% of its
cells fall in Level 1 and 62\% in Level 2, so the disagreement is not an
artefact of ambiguously shallow water, and imposing a minimum depth on
map B lowers its CSI monotonically (0.64 at 0 m, 0.62 at 0.3 m, 0.59 at
0.5 m) because recall falls faster than precision rises. The remaining
2.9 km\ensuremath{^{2}} that only B inundates is thin water at the edge of the
interpolated surface (median depth 0.18 m, 80\% of cells below 0.3 m).
The predicted severity levels are informative about the reference (Fig.
12d): 90\% of the cells predicted at Level 3 (\textgreater1.0 m) are
flooded in the reference, against 54\% at Level 2, 48\% at Level 1 and
13\% of the cells that B leaves dry. The Neuse River reach between
Kinston and New Bern (29 HWMs, not shown) gives the same picture (CSI
0.91 between the maps, 0.73 for both against the reference;
leave-one-out RMSE 1.10 m versus 0.69 m). The agreement between the two
maps cannot, however, be read as evidence that the predictions are
accurate. Replacing the predicted water surface by a constant depth of
0.50 m at the same 28 marks, the median depth of the training marks,
gives a map that agrees with the surveyed one at CSI 0.86, against 0.85
for the predictions, and the same result is obtained in nine further
reaches. The leave-one-out test explains why: interpolating the surveyed
water surface from the remaining 27 marks already has an RMSE of 1.25 m
at a held-out mark, whereas the point prediction error is 0.37 m, so the
error of a map built from 28 marks over an 84 km\ensuremath{^{2}} hull is dominated by
the interpolation and by the lidar terrain, not by the points. At this
density of marks, depth grids of the kind FEMA and the USGS produce are
insensitive to point-level accuracy below about one metre, and the
experiment therefore bounds what any point predictor, including ours,
can contribute to such a product. An operational product would require
the prediction points to be chosen before the event, for example along
the river network, and cross-section-based interpolation as used by the
USGS; whether a denser set of prediction points would make point
accuracy matter is left to future work.

\noindent\makebox[\linewidth][c]{\includegraphics[width=6.5in,height=5.10972in]{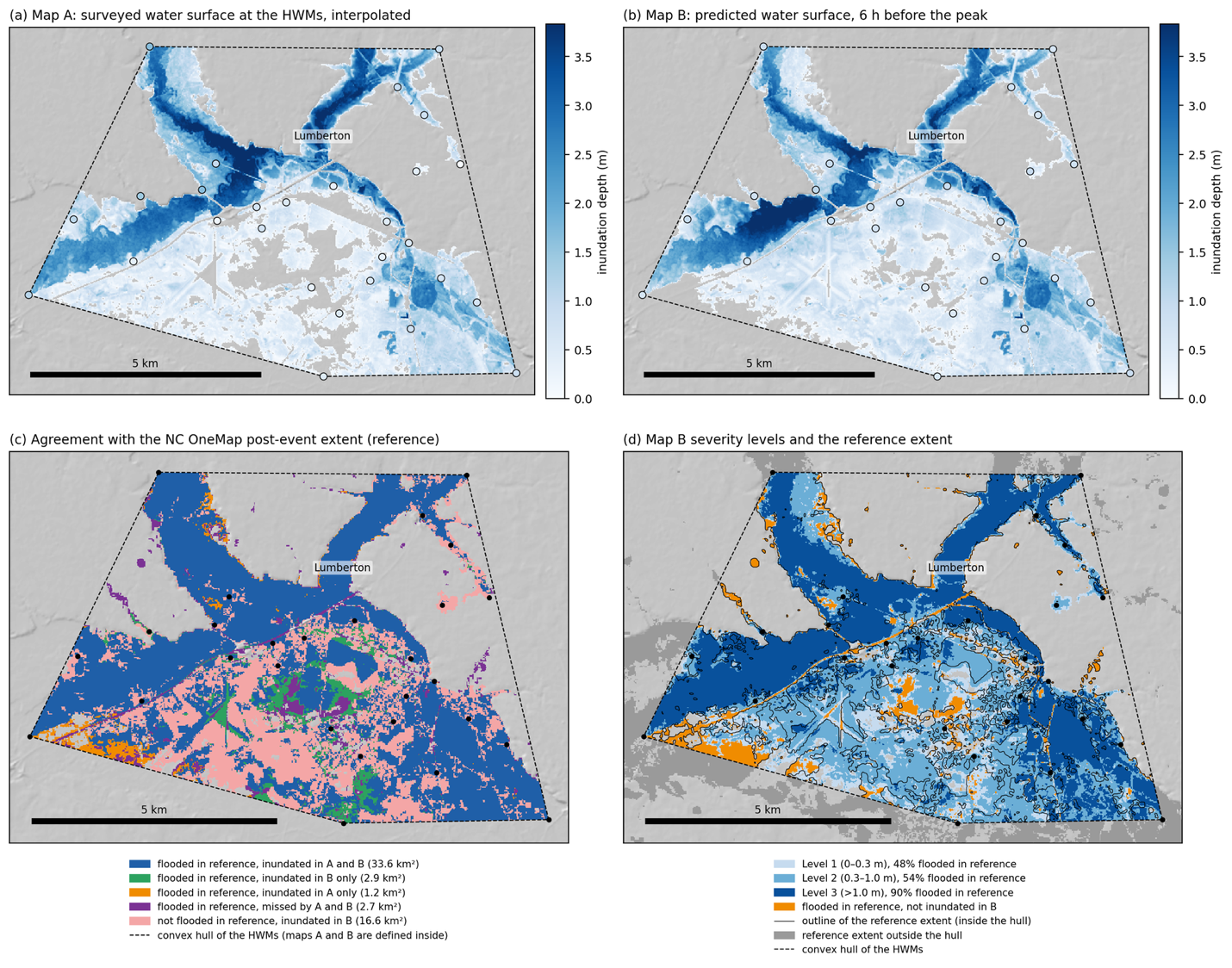}}

{\small
Fig. 12. Depth maps of Hurricane Florence on the Lumber River at
Lumberton built from point water-surface elevations at 28 HWMs (dots;
dashed line: convex hull of the marks, inside which the maps are
defined; background: 3DEP hillshade). (a) Map A: surveyed peak elevation
at the HWMs, interpolated (linear TIN) and minus the 3DEP 10 m DEM; dots
coloured by surveyed depth. (b) Map B: the same with the predicted water
surface (3DEP ground elevation at the mark plus the depth predicted 6 h
before the peak); dots coloured by predicted depth. (c) Agreement with
the NC OneMap post-event extent, taken as the reference: cells flooded
in the reference and inundated in both maps (blue), in B only (green),
in A only (orange) or in neither (purple), and cells inundated in B but
not flooded in the reference (pink); areas in km\ensuremath{^{2}}. (d) Severity levels
of map B (Level 1: 0--0.3 m; Level 2: 0.3--1.0 m; Level 3:
\textgreater1.0 m) with the outline of the reference extent inside the
hull, the reference extent outside the hull (grey) and the reference
flooded cells that B leaves dry (orange); the legend gives the share of
each level that is flooded in the reference.
\par}

\hypertarget{discussion}{%
\section{Discussion}\label{discussion}}

\hypertarget{comparison-with-existing-flood-depth-estimates}{%
\subsection{Comparison with existing flood depth estimates}\label{comparison-with-existing-flood-depth-estimates}}

Global flood models are validated against different quantities from
ours, which limits what a comparison can establish. For a 90-m global
compound flood model, \citet{rogers2025} report a water-surface
elevation RMSE of 2.08 m against FEMA 100-year base flood elevations and
1.26 m against coastal and lake gauges; against USGS high-water marks
they report results for the pluvial component in two events only, with
RMSEs of 1.47 m (July 2023, north-eastern United States) and 1.04 m
(Hurricane Ida, 2021). These are errors in water-surface elevation, so
the quantity in our results that corresponds to them is the
peak-elevation error rather than the depth error. On the event-disjoint
split our peak-elevation RMSE at the surveyed marks is 1.49 m, of the
same order, although our elevation is anchored on the lidar ground
elevation at each mark whereas theirs is derived from the global terrain
of the model. Our depth errors cannot be set against these figures, and
we make no claim of an advantage over \citet{rogers2025}; the
comparison serves only to place the magnitude of the errors in context.

Severity levels can be obtained either from the auxiliary
classification head or by thresholding the predicted depth at the level
boundaries of Table 1 (0.3 m and 1.0 m). The two give the same skill
(macro-F1 0.642 ± 0.011 versus 0.641 ± 0.003; micro-F1 0.727 ± 0.011
versus 0.731 ± 0.003; Fig. 13), so the classification head adds no
information beyond the predicted depth, which supports treating depth as
the primary output and deriving severity from it. The depth MAE of 0.26
m at the 6-h lead time is below the width of the lowest level (0.3 m),
although boundary cases near the thresholds remain uncertain, and a
four-class scheme with a separate level above 2.0 m could not be learned
reliably.

\noindent\makebox[\linewidth][c]{%
\includegraphics[width=0.49\linewidth]{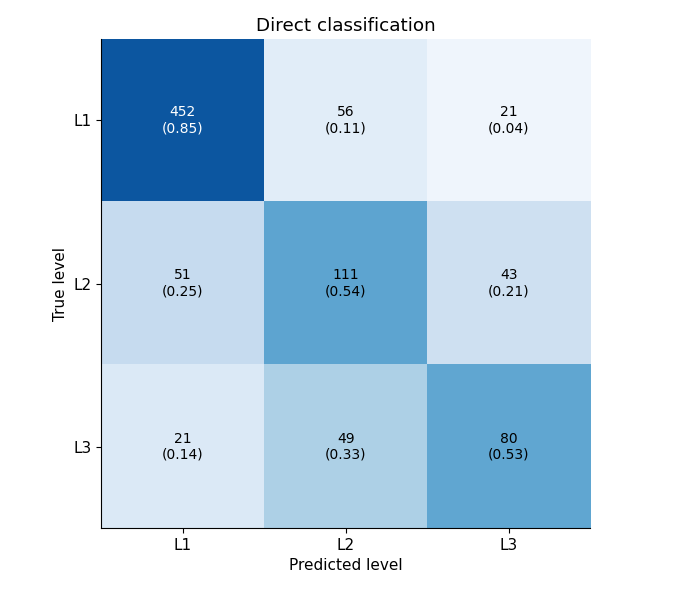}\hfill\includegraphics[width=0.49\linewidth]{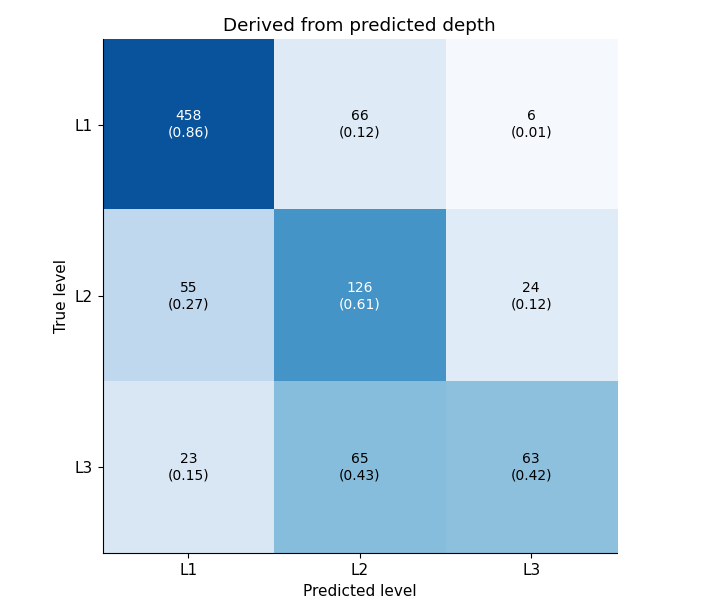}}

{\small
Fig. 13. Severity levels from the auxiliary classification head
and from the predicted depth. (a) Confusion matrix of the classification
head. (b) Confusion matrix of the levels obtained by thresholding the
predicted depth at 0.3 m and 1.0 m. 6-h lead time; counts averaged over
three seeds.
\par}

\hypertarget{implicit-correction-of-dem-error}{%
\subsection{Implicit correction of DEM error}\label{implicit-correction-of-dem-error}}

Extent-based depth estimation computes depth as the difference between a
water surface and the DEM, so the vertical error of the DEM passes into
the depth unchanged. The model studied here receives the same GLO-30
surface model as its only terrain input, but it is supervised by depths
surveyed in the field, so any systematic difference between the GLO-30
surface and the true ground at the marks, from canopy, buildings or the
interferometric error of the product, can be learned and compensated.
The data allow this to be tested directly, because the surveyed peak
water-surface elevation is known at every mark. At the 567 surveyed
HWMs of the random test set, differencing the surveyed water-surface
elevation with GLO-30 at the mark, which is the depth an extent-based
method would return given a perfect water surface, gives a depth MAE of
0.93 m (GLO-30 converted to NAVD88, the datum of the marks, as for the
model input; negative depths set to zero). With the 3DEP lidar ground
instead of GLO-30 the differenced depth has an MAE of 0.51 m, and the
model reaches 0.39 m (Table 8). The differenced depth deteriorates as
the local GLO-30 error grows, from an MAE of 0.65 m where GLO-30 lies
within 1 m of 3DEP (183 marks) to 1.10 m and 1.03 m where it deviates by
1--3 m and by more than 3 m (207 and 177 marks), whereas the model error
stays at 0.37, 0.37 and 0.44 m. Under tree canopy (NLCD forest and woody
wetland, 99 marks), where GLO-30 lies a median 5.3 m above the lidar
ground, the model error is 0.45 m against 1.08 m for the differenced
depth, and in built-up areas (299 marks) it is 0.36 m against 0.87 m.
Within the sampled events the depth learned from surveyed depths is
therefore largely insensitive to the vertical error of the DEM it is
given. This correction does not transfer to unseen events. On the event
split the model error at the 575 marks (0.57 m) is lower than that of
the differenced depth (1.04 m), but equal to that of a constant
prediction of the training median depth (0.57 m), and the same holds in
every DEM-error and land-cover stratum; on new events the model
therefore avoids the DEM error by not relying on absolute terrain rather
than by correcting it, and, as Section 4.3 shows, it does not resolve
depth at those marks at all. The implicit correction demonstrated here
is thus a property of the model within the events and regions it has
seen, not a general capability. Two caveats apply. The correction is
learned from marks in surveyed regions and is therefore expected to hold
where the GLO-30 error has the same causes; and the residual
peak-elevation task, whose anchor is the 3DEP ground elevation, is not
evidence for it, because its labels are already free of GLO-30 error.

\begin{table}[!htbp]
\centering
\small
\caption{Depth error at the 567 surveyed HWMs of the random test set, by local GLO-30 error and land cover. Differenced depths subtract GLO-30 (converted to NAVD88, the datum of the marks) or the 3DEP lidar ground at the mark from the surveyed water-surface elevation (WSE), with negative depths set to zero, i.e. the depth an extent-based method would return given a perfect water surface; TCM-Former predicts depth at the 6-h lead time (mean over three seeds). Land cover from NLCD 2021 (tree canopy: forest and woody wetland; built-up: developed, low to high intensity).}\label{tab:8}
\setlength{\tabcolsep}{3pt}
\begin{tabular*}{\textwidth}{@{\extracolsep{\fill}}lccccc@{}}
\toprule
Subset of test HWMs & $n$ & Median & \multicolumn{3}{c}{Depth MAE (m)} \\
\cmidrule(l){4-6}
(random split) & & $|$GLO-30 $-$ 3DEP$|$ (m) & WSE $-$ GLO-30 & WSE $-$ 3DEP & TCM-Former \\
\midrule
All HWMs & 567 & 1.66 & 0.93 & 0.51 & 0.39 \\
\midrule
$|$GLO-30 $-$ 3DEP$|$ $<$ 1 m & 183 & 0.55 & 0.65 & 0.39 & 0.37 \\
$|$GLO-30 $-$ 3DEP$|$ 1--3 m & 207 & 1.68 & 1.10 & 0.54 & 0.37 \\
$|$GLO-30 $-$ 3DEP$|$ $>$ 3 m & 177 & 6.47 & 1.03 & 0.61 & 0.44 \\
\midrule
Tree canopy & 99 & 5.30 & 1.08 & 0.69 & 0.45 \\
Built-up & 299 & 1.28 & 0.87 & 0.44 & 0.36 \\
\bottomrule
\end{tabular*}
\end{table}

The auxiliary peak-elevation task is the one most constrained by the
terrain input. The global RMSE of GLO-30 is approximately 4 m \citep{simard2024}, which exceeds most flood depths in our dataset; because
GLO-30 is a surface model, buildings and canopy also bias ground
elevation upward. The residual target itself is anchored on the 3DEP
ground elevation at the mark (Section 3.3), so this error does not enter
the labels; it enters through the only terrain input of the model, the
GLO-30 patch, from which the network has to infer the local ground level
under canopy and buildings. This is consistent with residual errors that
do not improve with spatial proximity (Table 6). Depth and severity are
less affected because they are supervised by surveyed depths and can be
learned from relative terrain and land-surface patterns rather than
absolute elevation. Unlike hydrodynamic models, in which DEM error
propagates through the governing equations \citep{saksena2015},
the learned representation can partly absorb systematic terrain bias
\citep{bentivoglio2025}.

Two practical implications follow. Operational peak-elevation products
require bare-earth terrain such as lidar-based 3DEP or forest- and
building-corrected global DEMs such as FABDEM \citep{hawker2022},
whereas depth and severity are already the more transferable outputs of
the framework. Because HWM campaigns report elevation more often than
depth, pairing elevation-only HWMs with lidar DEMs is a direct route to
more training labels.

\hypertarget{choice-of-the-meteorological-descriptor-and-drivers-of-predicted-flood-depth}{%
\subsection{Choice of the meteorological descriptor and drivers of predicted flood depth}\label{choice-of-the-meteorological-descriptor-and-drivers-of-predicted-flood-depth}}

We also tested whether learned temporal encoders extract more
flood-relevant information from the forcing than the hand-crafted
three-tier descriptor (Section 3.2; Table 3). Replacing the 58
descriptors per window with 140 TabPFN-TS forecast features per window
gives a depth MAE of 0.258 m and a depth RMSE of 0.457 m, against 0.257
and 0.468 m for the three-tier descriptor, and an LSTM over the raw
hourly series gives 0.267 and 0.461 m. For neither encoder does any of
the six metrics differ significantly from the three-tier descriptor
(Welch\textquotesingle s t-test over three seeds, p \textgreater{} 0.1),
and the auxiliary tasks are equally unaffected (residual elevation MAE
0.541 and 0.545 m versus 0.530 m; macro-F1 0.647 and 0.653 versus
0.642). When meteorology is the only input, the LSTM encoder scores
higher on average on the auxiliary severity task (macro-F1 0.615 versus
0.559), but the difference is not significant (p = 0.14) and disappears
once terrain and embeddings are added. The learned encoders therefore do
not extract information beyond what the descriptor captures: for this
task the decisive information, how much water reached the site and how
recently, is carried by accumulation totals, antecedent indices and
storm-state indicators that can be written down explicitly. The learned
features, the hidden state of the LSTM and the forecasts of the
pretrained TabPFN-TS model, also have no direct physical meaning, and
TabPFN-TS needs tens of seconds of model inference per sample to produce
them, whereas the descriptor is computed analytically. We therefore
retain the three-tier descriptor: every one of its features has an
explicit physical definition, so the contribution of each feature to the
predicted depth can be quantified with Shapley-value attribution and
read directly in terms of hydrometeorological processes.

Shapley-value attribution \citep{lundberg2017} for the 6-h
model (Fig. 14) was computed with Expected Gradients for the predicted
flood depth in metres, over the 58 × 6 meteorological inputs of the 885
test samples, keeping each sample\textquotesingle s terrain and
embedding inputs fixed, and averaged over three seeds. The predicted
depth is driven mainly by the water supplied to the site. The three
top-ranked descriptors are Tier-3 water-balance terms of the most recent
window (6--24 h before the peak): cumulative net water input, the
antecedent precipitation index and cumulative precipitation, each
changing the predicted depth by about 0.03 m on average and together
accounting for 25\% of the total attribution; together with the window
totals of precipitation, liquid rainfall and net water input, the
precipitation-amount and antecedent-wetness descriptors account for
45\%. The second major group (31\%) comprises storm-state indicators,
namely the maximum wind gust, the maximum WMO weather code, cloud cover
and wind speed, which separate organized storm systems, such as the
tropical cyclones that dominate the dataset, from fair-weather
conditions. Rainfall-intensity statistics such as the maximum 3-h
accumulation contribute less (6\%), as do pressure (6\%) and temperature
and humidity descriptors (7\%). Attribution is concentrated near the
peak: the window 6--24 h before the peak accounts for 66\% of the total
and the preceding day for 13\%, whereas the three windows more than 72 h
before the peak contribute 14\% together. Feature rankings are stable
across seeds (Spearman \ensuremath{\rho} = 0.96--0.97 between seeds) and similar for
HWMs and rain-free samples. The attributions of the depth and of the
auxiliary severity logit are nearly collinear within each sample (median
\textbar r\textbar{} = 0.998), so the network routes the forcing through
essentially one learned direction, an index of how much water the storm
delivered, on which depth and severity both depend. Together with the
modality attribution in Section 4.2, this indicates that recent water
input and storm conditions carry most of the depth signal, while the
spatial inputs reduce the remaining error at the surveyed marks by about
a fifth.

\noindent\makebox[\linewidth][c]{\includegraphics[width=6in,height=4.34607in]{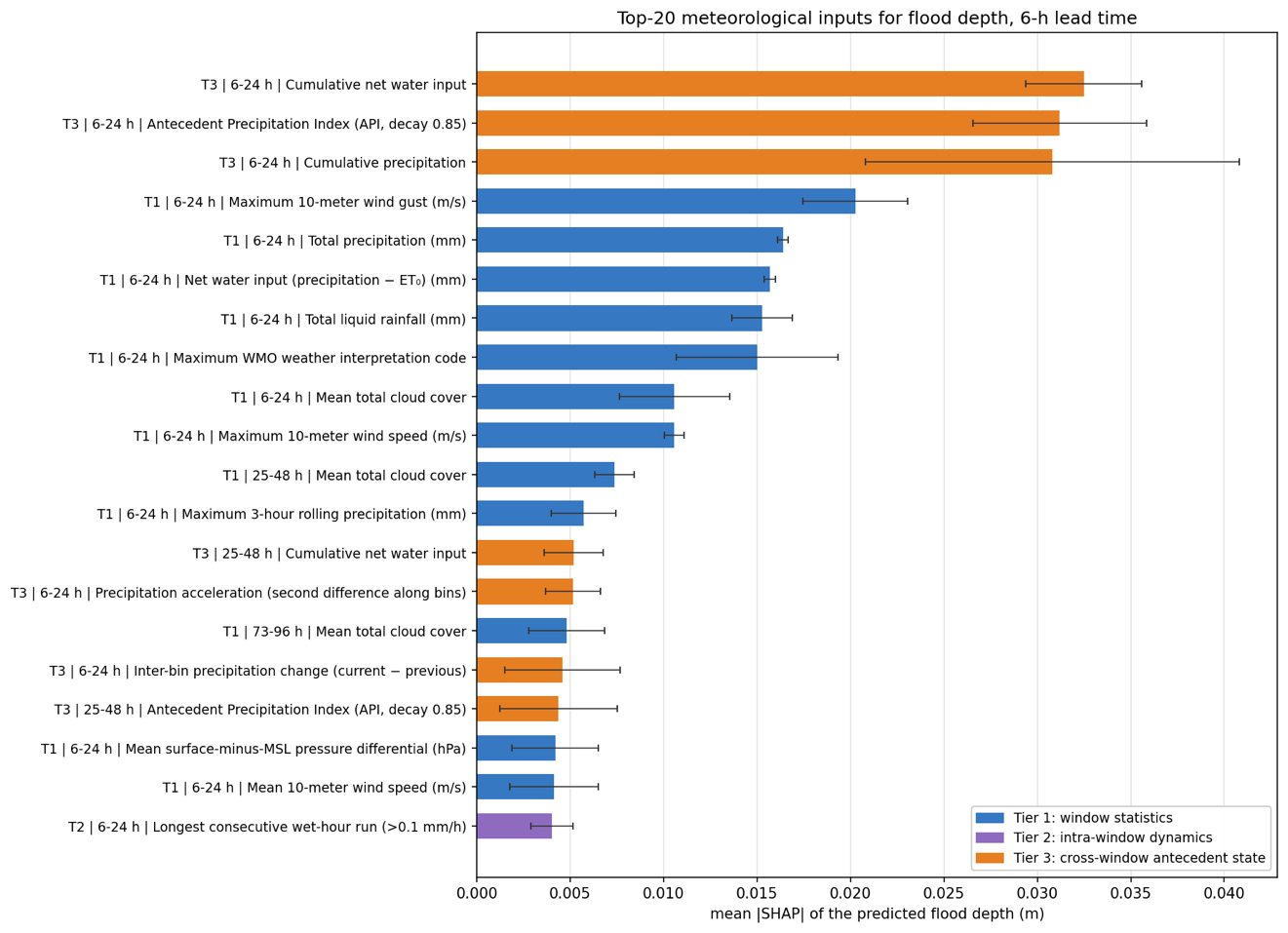}}

{\small
Fig. 14. Top 20 meteorological inputs ranked by mean absolute SHAP
value of the predicted flood depth at the 6-hour lead time (Expected
Gradients; mean over 885 test samples and three seeds; error bars: s.d.
over seeds). Labels give the tier, the 24-h window (hours before the
peak; the most recent window spans 6--24 h because of the lead-time
mask) and the feature. High SHAP magnitude indicates strong model
sensitivity but does not necessarily imply a positive net contribution
to predictive performance.
\par}

\hypertarget{limitations-and-outlook}{%
\subsection{Limitations and outlook}\label{limitations-and-outlook}}

Four limitations bound the present results. First, lead times are
referenced to the peak time of the matched gauge within 10 km rather
than to the local inundation peak at each HWM; The peak time is moreover
taken from the matched sensor rather than predicted, so the skill
reported here is conditional on the timing of the peak. An operational
system would obtain that timing from a stage or discharge forecast, or
run the model on a rolling window and issue the peak depth implied by
each candidate timing; its skill would then also depend on the accuracy
of the timing.the two need not coincide \citep{allen2018,yang2016}, which is why lead times shorter than 6 h were not evaluated
(Section 4.1). The size of the offset is not quantified here, and the
nominal lead times should be read as referring to the sensor peak rather
than to the local one. Second, and most importantly, depth skill does
not transfer. On unseen events the model still distinguishes flooded
from rain-free conditions, but among the surveyed marks the predicted
depths occupy a narrow band (slope 0.06, coefficient of determination
0.03) and carry no advantage over a constant prediction (Section 4.3).
One likely cause is the distribution of the labels rather than the
architecture: 4650 of the 8874 samples, 52\%, have a depth of exactly
zero, counting the 3185 rain-free samples and the 1465 marks found at
ground level, and a squared-error objective on such a target is
minimised by predicting zeros accurately and shrinking everything else
towards the mean. The under-dispersion of the predictions on the random
split itself, where the regression slope is only 0.36, is consistent
with this. A two-stage formulation, in which the classification head
separates flooded from dry sites and the depth head is trained only on
the marks with non-zero depth, is the natural test of whether the
failure to transfer is intrinsic to the task or an artefact of the
objective, and we have not yet carried it out. The training events are
also concentrated in the humid eastern United States, and skill
decreases for unseen events and spatially disjoint sub-regions (Tables 5
and 6), so the model currently behaves as an interpolator within
sampled hydroclimates. Because samples are split at random by row,
HWMs from the same event and neighbourhood occur in both training and
test sets, so part of the reported skill may reflect geographic
memorization; the event- and space-disjoint evaluations in Section 4.3
quantify the transferable part. Third, the rain-free samples contrast
each site with itself under no rain; samples of the same site under
heavy rain without inundation cannot be constructed from HWM surveys,
because the absence of a mark does not establish the absence of
flooding. The supervision therefore does not teach the model where water
does not go under rain, which confines its use to sites comparable to
the surveyed ones; this limitation stems from the availability of
observed labels and is shared by every dataset built from field
surveys. Fourth, the embeddings are annual and the DEM is static, so
event-specific surface states such as soil moisture or pre-event
inundation enter only through meteorology. Event-time EO observations,
such as SAR-derived soil moisture or pre-peak Sentinel-1 water extent,
are the natural next inputs, together with lidar or InSAR terrain and
international HWM archives.

\hypertarget{conclusion}{%
\section{Conclusion}\label{conclusion}}

This study asked whether flood peak depth can be learned from field
observations and predicted before the peak from spaceborne terrain,
geospatial foundation-model embeddings and meteorological forcing. Using
5689 USGS high-water marks from 39 events, with sensor-derived peak
times and site-matched rain-free windows, a cross-modal fusion network
predicts depth at surveyed marks with an MAE of 0.39 m six hours
before the peak and 0.41 m a day before, and 0.57 m on eleven events it
has never seen. Three findings follow. First, deep-learning flood depth
prediction does not depend on simulated labels: trained on observations
alone, the model reaches an accuracy below the 0.3--1.0 m band that
separates nuisance from damaging flooding. Second, supervision by
surveyed depth corrects the vertical error of the spaceborne DEM
implicitly, reducing the depth error at the surveyed marks from 0.93
m for water levels differenced with GLO-30 to 0.39 m, which extent-based
depth estimation cannot do; on unseen events, however, the error at the
marks (0.57 m) is no lower than that of a constant prediction. Third,
predictions issued before the peak, interpolated with the procedure FEMA
and the USGS apply to surveyed marks, give a depth map of a held-out
hurricane equivalent to the post-event map, but a constant water surface
at the same marks does so equally well, which shows that such grids are
governed by the interpolation and the terrain rather than by the
accuracy of the points. Meteorological forcing carries most of the
skill, and terrain and embeddings, largely interchangeable, reduce the
depth error at the surveyed marks by about a fifth. Expanding HWM-based
benchmarks, adopting bare-earth terrain and adding event-time EO
observations are the most direct paths toward observation-supervised,
EO-driven flood depth forecasting and mapping. The principal limitation
is that depth skill does not transfer: on unseen events the model still
separates flooded from rain-free conditions, but among the surveyed
marks it does not resolve depth at all, and the strongly under-dispersed
predictions suggest that the zero-inflation of the labels, rather than
the architecture, may be responsible. Lead times are moreover defined
relative to an observed peak.

\appendix
\hypertarget{appendix-a.-the-time-series-factors-obtained-from-open-meteo-api}{%
\section{The time-series factors obtained from Open-Meteo API}\label{appendix-a.-the-time-series-factors-obtained-from-open-meteo-api}}

\setcounter{table}{0}\renewcommand{\thetable}{A\arabic{table}}
{\small
\begin{longtable}{@{}>{\raggedright\arraybackslash}p{5.2cm}>{\raggedright\arraybackslash}p{\dimexpr\textwidth-5.2cm-2\tabcolsep\relax}@{}}
\caption{The used time-series factors obtained from Open-Meteo API}\label{tab:A1}\\
\toprule
Variable & Description \\
\midrule
\endfirsthead
\toprule
Variable & Description \\
\midrule
\endhead
\bottomrule
\endlastfoot
\texttt{time} & Hourly flood peak timestamp string \\
\texttt{temperature\_2m} & Air temperature at 2 meters above ground \\
\texttt{relative\_humidity\_2m} & Relative humidity at 2 meters above ground \\
\texttt{dewpoint\_2m} & Dew point temperature at 2 meters above ground \\
\texttt{apparent\_temperature} & ``Feels like'' temperature (derived from temperature, wind, humidity, radiation, etc.) \\
\texttt{precipitation} & Total precipitation (commonly rain + snow water equivalent) \\
\texttt{rain} & Rainfall component of precipitation \\
\texttt{snowfall} & Snowfall (definition/units depend on requested settings) \\
\texttt{weather\_code} & WMO weather interpretation code (categorical weather condition) \\
\texttt{pressure\_msl} & Mean sea level pressure \\
\texttt{surface\_pressure} & Surface pressure at the location \\
\texttt{cloud\_cover} & Total cloud cover \\
\texttt{cloud\_cover\_low} & Low‑level cloud cover \\
\texttt{et0\_fao\_evapotranspiration} & Reference evapotranspiration (FAO‑56 ET0) \\
\texttt{vapour\_pressure\_deficit} & Vapour pressure deficit (VPD; a dryness indicator) \\
\texttt{wind\_speed\_10m} & Wind speed at 10 meters above ground \\
\texttt{wind\_gusts\_10m} & Wind gust speed at 10 meters above ground \\
\texttt{cloud\_cover\_mid} & Mid-level cloud cover \\
\texttt{cloud\_cover\_high} & High-level cloud cover \\
\texttt{wind\_direction\_10m} & Wind direction at 10 meters above ground \\
\texttt{wind\_speed\_100m} & Wind speed at 100 meters above ground \\
\texttt{hours\_before\_peak} & Time offset in hours before the flood peak time \\
\texttt{days\_before\_peak} & Time offset in days before the flood peak time \\
\end{longtable}
}

\hypertarget{appendix-b.-detailed-description-of-physics-motivated-hydrometeorological-domain-features-tier-1}{%
\section{Detailed description of physics-motivated hydrometeorological domain features (Tier 1)}\label{appendix-b.-detailed-description-of-physics-motivated-hydrometeorological-domain-features-tier-1}}

Each 144-hour lookback window is partitioned into six
contiguous 24-hour bins. Within every bin, the physically defined
features of the three tiers are concatenated to form a 58-dimensional
tabular descriptor (indices 0--34 and 47--69; indices 35--46 hold the
TabPFN-TS diagnostics of Appendix C, which are not part of the
descriptor); this appendix documents \textbf{Tier 1}, which occupies
the 35 dimensions at window indices 0--34 (0-based). The five groups
below characterize the physical forcing mechanisms most directly linked
to flood generation: precipitation loading, phase and intensity
structure, atmospheric moisture availability, synoptic circulation
strength and trend, and the surface energy balance governing pre-event
soil moisture state. All features are computed analytically without
model inference, as are those of Tiers 2 and 3.

\textbf{Group 1 --- Precipitation Intensity and Structure (11 features,
indices 0--10)}

This group characterizes the magnitude, variability, and temporal
organization of rainfall. The eleven features are:

\textbf{(1) \emph{Total precipitation --- }}total rainfall accumulation
(mm) over the 24-hour bin, representing the overall precipitation
loading.

\textbf{(2) \emph{Maximum hourly precipitation --- }}peak hourly
precipitation rate (mm\,h\ensuremath{^{-1}}), capturing the highest instantaneous
rainfall intensity within the bin.

\textbf{(3) \emph{Mean hourly precipitation --- }}mean hourly
precipitation rate (mm\,h\ensuremath{^{-1}}) over all hours of the bin.

\textbf{(4) \emph{Standard deviation of hourly precipitation ---
}}standard deviation of the hourly precipitation series (mm\,h\ensuremath{^{-1}}),
encoding within-bin rainfall variability.

\textbf{(5) \emph{Total liquid rainfall --- }}total liquid-phase
rainfall accumulation (mm), excluding solid-phase snowfall
contributions.

\textbf{(6) \emph{Total snowfall water equivalent --- }}total snowfall
accumulation expressed as liquid water equivalent (mm), relevant to
snowmelt-driven flood precursors.

\textbf{(7) \emph{Wet-hour count --- }}number of hours within the bin
for which precipitation exceeds 0.1\,mm\,h\ensuremath{^{-1}}, encoding the duration of
active rainfall.

\textbf{(8) \emph{Heavy-rain hour count --- }}number of hours with
precipitation exceeding 5.0\,mm\,h\ensuremath{^{-1}}, indicative of convective or intense
frontal rainfall.

\textbf{(9) \emph{Mean precipitation intensity per wet hour --- }}mean
precipitation intensity conditional on wet hours (total precipitation
divided by wet-hour count, mm\,h\ensuremath{^{-1}}), which decouples rainfall duration
from intensity.

\textbf{(10) \emph{Maximum 3-hour rolling precipitation --- }}maximum
value of a 3-hour sliding-window accumulation (mm), capturing the most
intense short-duration burst within the bin and particularly relevant to
flash-flood triggering.

\textbf{(11) \emph{Liquid-rain fraction of total precipitation ---
}}ratio of liquid rainfall to total precipitation (dimensionless, 0--1),
distinguishing rainfall-driven from snowmelt-driven flood precursors.

\textbf{Group 2 --- Thermodynamic State (8 features, indices 11--18)}

Temperature and humidity fields govern evapotranspiration demand,
snowmelt potential, and near-surface atmospheric stability. The eight
features are:

\textbf{(12) \emph{Mean 2-meter air temperature --- }}mean 2-m air
temperature (°C) over the 24-hour bin.

\textbf{(13) \emph{Minimum 2-meter air temperature --- }}minimum 2-m air
temperature (°C) within the bin, relevant to freezing-level diagnosis
and phase partitioning of precipitation.

\textbf{(14) \emph{Diurnal range of 2-meter air temperature ---
}}diurnal temperature range (maximum minus minimum, °C), encoding the
magnitude of the daily thermal cycle and convective instability
potential.

\textbf{(15) \emph{Linear trend of 2-meter air temperature ---
}}least-squares linear slope of the hourly 2-m temperature series
against the temporal index (°C\,h\ensuremath{^{-1}}), indicating whether the bin is
warming or cooling.

\textbf{(16) \emph{Mean 2-meter relative humidity --- }}mean 2-m
relative humidity (\%) over the bin, encoding the bulk near-surface
moisture content.

\textbf{(17) \emph{Minimum 2-meter relative humidity --- }}minimum 2-m
relative humidity (\%) within the bin, a proxy for near-surface dryness
and evaporative demand.

\textbf{(18) \emph{Mean 2-meter dew-point temperature --- }}mean 2-m
dew-point temperature (°C), providing an absolute measure of atmospheric
moisture content independent of temperature.

\textbf{(19) \emph{Mean vapour pressure deficit --- }}mean vapour
pressure deficit (VPD, kPa) over the bin, quantifying atmospheric drying
power and the upper bound on potential evapotranspiration.

\textbf{Group 3 --- Synoptic Pressure Dynamics (4 features, indices
19--22)}

Mean sea-level pressure (MSLP) and its temporal evolution provide direct
indicators of synoptic-scale forcing. The four features are:

\textbf{(20) \emph{Mean sea-level pressure --- }}mean MSLP (hPa) over
the 24-hour bin, characterising the prevailing large-scale pressure
environment.

\textbf{(21) \emph{Minimum sea-level pressure --- }}minimum MSLP (hPa)
within the bin, associated with the deepest low-pressure system passing
through the analysis period.

\textbf{(22) \emph{Linear trend of sea-level pressure ---
}}least-squares linear slope of the hourly MSLP series (hPa\,h\ensuremath{^{-1}});
negative values indicate deepening cyclonic development conducive to
heavy rainfall.

\textbf{(23) \emph{Sea-level pressure range --- }}intra-bin MSLP range
(maximum minus minimum, hPa), distinguishing the regular semi-diurnal
atmospheric tide from synoptic-scale pressure falls.

\textbf{Group 4 --- Wind Field (4 features, indices 23--26)}

Wind speed mediates moisture transport and surface roughness effects on
precipitation distribution. The four features are:

\textbf{(24) \emph{Mean 10-meter wind speed --- }}mean 10-m wind speed
(m\,s\ensuremath{^{-1}}) over the bin, encoding the prevailing advective flow.

\textbf{(25) \emph{Maximum 10-meter wind speed --- }}maximum 10-m wind
speed (m\,s\ensuremath{^{-1}}) within the bin.

\textbf{(26) \emph{Maximum 10-meter wind gust --- }}maximum 10-m wind
gust speed (m\,s\ensuremath{^{-1}}), particularly relevant to convective storm
identification.

\textbf{(27) \emph{Linear trend of 10-meter wind speed ---
}}least-squares linear slope of the hourly 10-m wind speed series
(m\,s\ensuremath{^{-1}}\,h\ensuremath{^{-1}}), indicating whether the synoptic flow is strengthening or
weakening.

\textbf{Group 5 --- Surface Energy Balance and Cloud Regime (8 features,
indices 27--34)}

The remaining features characterize the radiation environment,
evapotranspiration demand, and cloud regime governing surface moisture
state. The eight features are:

\textbf{(28) \emph{Mean total cloud cover --- }}mean total cloud cover
fraction (0--1) over the bin.

\textbf{(29) \emph{Linear trend of total cloud cover --- }}least-squares
linear slope of the hourly total cloud cover series, encoding the onset
or clearance of cloud systems within the bin.

\textbf{(30) \emph{Total reference evapotranspiration FAO-56 ET\ensuremath{_{0}} ---
}}accumulated FAO-56 reference evapotranspiration ET\ensuremath{_{0}} (mm) over the bin,
computed via the Penman--Monteith formulation and representing the
potential atmospheric water demand.

\textbf{(31) \emph{Net water input --- }}net water input (P−ET\ensuremath{_{0}}, mm),
computed as the difference between total precipitation and ET\ensuremath{_{0}}
accumulation, approximating the available water surplus driving
infiltration excess and overland flow.

\textbf{(32) \emph{Mean apparent-minus-actual 2-meter temperature excess
--- }}mean difference between apparent temperature and actual 2-m air
temperature (°C), encoding non-linear atmospheric stress effects due to
wind and humidity.

\textbf{(33) \emph{Mean surface-minus-sea-level pressure differential
--- }}mean difference between surface pressure and MSLP (hPa),
reflecting orographic and boundary-layer effects on the local pressure
field.

\textbf{(34) \emph{Mean low-level cloud cover --- }}mean low-level cloud
cover fraction (0--1), associated with stratiform precipitation regimes.

\textbf{(35) \emph{Maximum WMO weather interpretation code --- }}maximum
WMO weather code observed within the bin, providing a categorical
summary of the most severe meteorological phenomenon (e.g.,
thunderstorm, heavy snow, blizzard).

Together, the 35 Tier 1 features encode the physical forcing mechanisms
most directly linked to flood generation. All features are computed
analytically without model inference, as are those of Tiers 2 and
3.

\hypertarget{appendix-c.-detailed-description-of-intra-window-meteorological-dynamics-and-coupling-features-tier-2}{%
\section{Detailed description of intra-window meteorological dynamics and coupling features (Tier 2)}\label{appendix-c.-detailed-description-of-intra-window-meteorological-dynamics-and-coupling-features-tier-2}}

This appendix documents Tier 2, whose 13 physically defined features
occupy window indices 47--59 (0-based) within each 24-hour bin. Tier 2
features capture dynamic intra-window behaviour (temporal
predictability, meteorological variable coupling, signal volatility, and
extreme-event structure) that complements the scalar summary statistics
of Tier 1. The 13 features are generated by Processes 2--4, which
compute explicit analytic statistics directly from the hourly Open-Meteo
time series. Process 1, which applies the TabPFNTimeSeriesPredictor
(TabPFN-TS) in univariate mode to three meteorological series (indices
35--46), is documented for completeness but is not part of the
three-tier descriptor; the TabPFN-TS encoder compared in Sections 4.1
and 5.3 is a separate, wider feature set described in Section 3.2.

\textbf{Process 1: TabPFN-TS Univariate Predictability Features (12
features, indices 35--46; not part of the three-tier descriptor)}

Process 1 applies ``TabPFNTimeSeriesPredictor'' independently to the
hourly precipitation, mean sea-level pressure, and 2-m air temperature
series of each 24-hour bin. For each variable, four scalar diagnostics
are extracted: a short-horizon forecast value, the trend along the
prediction trace, a backtest root-mean-square error (RMSE), and a
seasonal structure strength derived from the built-in sin/cos feature
channels of the ``FeatureTransformer''.

\emph{\textbf{P1.1 --- Precipitation sub-block (indices 35--38)}}

\textbf{(36) \emph{TabPFN-TS precipitation short-horizon forecast level
--- }}one-step / short-horizon precipitation forecast value at the
prediction side of the 24-hour bin, produced by running
``TabPFNTimeSeriesPredictor'' on the hourly precipitation series.

\textbf{(37) \emph{TabPFN-TS precipitation forecast trend --- }}change
of the short-horizon forecast along the model's own prediction horizon
(end minus start of the predicted trace), encoding the momentum of the
predicted precipitation.

\textbf{(38) \emph{In-bin backtest RMSE for precipitation ---
}}backtest-style RMSE evaluating the holdout tail of the bin against a
linear trend fitted to the first 80\% of the hourly precipitation
sequence, measuring departure from a simple linear baseline.

\textbf{(39) \emph{Precipitation seasonal-channel strength ---
}}seasonal structure strength for precipitation, computed as the mean
standard deviation of the built-in sin/cos seasonal feature channels
output by the ``FeatureTransformer'' module.

\emph{\textbf{P1.2 --- Mean sea-level pressure sub-block (indices
39--42)}}

\textbf{(40) \emph{TabPFN-TS mean sea-level pressure short-horizon
forecast level --- }}short-horizon TabPFN-TS forecast value for mean
sea-level pressure, applying the same pipeline as the precipitation
sub-block.

\textbf{(41) \emph{TabPFN-TS mean sea-level pressure forecast trend ---
}}forecast trend along the TabPFN-TS prediction run for MSL pressure
(end minus start of the predicted trace).

\textbf{(42) \emph{In-bin backtest RMSE for mean sea-level pressure ---
}}backtest RMSE for MSL pressure, evaluating the holdout tail against a
linear in-sample trend.

\textbf{(43) \emph{Mean sea-level pressure seasonal-channel strength ---
}}seasonal structure strength for MSL pressure, derived from the sin/cos
channel standard deviations of the ``FeatureTransformer''.

\emph{\textbf{P1.3 --- 2-meter temperature sub-block (indices 43--46)}}

\textbf{(44) \emph{TabPFN-TS 2-meter temperature short-horizon forecast
level --- }}short-horizon TabPFN-TS forecast value for 2-m air
temperature, using the same univariate pipeline as the other two
sub-blocks.

\textbf{(45) \emph{TabPFN-TS 2-meter temperature forecast trend ---
}}forecast trend along the TabPFN-TS prediction run for 2-m temperature
(end minus start of the predicted trace).

\textbf{(46) \emph{In-bin backtest RMSE for 2-meter temperature ---
}}backtest RMSE for 2-m temperature, evaluating the holdout tail against
a linear in-sample trend.

\textbf{(47) \emph{2-meter temperature seasonal-channel strength ---
}}seasonal structure strength for 2-m temperature, derived from the
sin/cos channel standard deviations of the ``FeatureTransformer''.

\textbf{Process 2: Multivariate Meteorological Coupling (6 features,
indices 47--52)}

Process 2 quantifies linear co-variation between pairs of meteorological
variables using Pearson correlation coefficients computed over the
hourly records within the bin. These coupling features are
physics-interpretable and are computed analytically without model
inference. Six variable pairs are evaluated:

\textbf{(48) \emph{Pearson correlation: precipitation vs mean
sea-level pressure --- }}Pearson r between hourly \emph{Total
precipitation} and \emph{Mean sea-level pressure}, providing a direct
indicator of synoptic cyclone-forced rainfall.

\textbf{(49) \emph{Pearson correlation: precipitation vs relative
humidity --- }}Pearson r between hourly \emph{Total precipitation} and
\emph{Mean 2-meter relative humidity}, encoding near-surface moisture
availability during rainfall events.

\textbf{(50) \emph{Pearson correlation: 2-meter temperature vs mean
sea-level pressure --- }}Pearson r between \emph{Mean 2-meter air
temperature} and \emph{Mean sea-level pressure}, reflecting the thermal
structure associated with synoptic-scale pressure systems.

\textbf{(51) \emph{Pearson correlation: precipitation vs 10-meter wind
speed --- }}Pearson r between \emph{Total precipitation} and \emph{Mean
10-meter wind speed}, sensitive to convective versus frontal
precipitation regimes.

\textbf{(52) \emph{Pearson correlation: precipitation vs cloud cover ---
}}Pearson r between \emph{Total precipitation} and \emph{Mean total
cloud cover}, characterising the temporal co-occurrence of cloud
radiative forcing and surface rainfall.

\textbf{(53) \emph{Pearson correlation: relative humidity vs mean
sea-level pressure --- }}Pearson r between \emph{Mean 2-meter relative
humidity} and \emph{Mean sea-level pressure}, encoding the
boundary-layer moisture response to synoptic pressure variations.

\textbf{Process 3: First-Difference Volatility (3 features, indices
53--55)}

Process 3 measures hour-to-hour variability by computing the standard
deviation of the first-difference series (\ensuremath{\sigma}(\ensuremath{\Delta}x\ensuremath{_{t}}) where \ensuremath{\Delta}x\ensuremath{_{t}}\,=\,x\ensuremath{_{t}}−x\ensuremath{_{t-1}}).
Three scalars are produced:

\textbf{(54) \emph{Sub-daily precipitation volatility --- }}standard
deviation of first differences of the hourly \emph{Total precipitation}
series (mm\,h\ensuremath{^{-1}}), quantifying temporal jitter and rainfall intermittency.

\textbf{(55) \emph{Sub-daily mean sea-level pressure volatility ---
}}standard deviation of first differences of the hourly \emph{Mean
sea-level pressure} series (hPa\,h\ensuremath{^{-1}}), encoding the rapidity of synoptic
pressure change.

\textbf{(56) \emph{Sub-daily 2-meter temperature volatility ---
}}standard deviation of first differences of the hourly \emph{Mean
2-meter air temperature} series (°C\,h\ensuremath{^{-1}}), capturing intra-bin thermal
instability associated with frontal passages or radiative cycles.

\textbf{Process 4: Extreme-Event and Distributional Structure (4
features, indices 56--59)}

Process 4 characterizes the tail behavior and temporal clustering of
precipitation. All four features are computed analytically:

\textbf{(57) \emph{Precipitation 2-sigma anomaly hour count --- }}number
of hours within the bin for which \emph{Total precipitation} exceeds the
bin mean plus two standard deviations (mean\,+\,2\ensuremath{\sigma}), identifying
statistically anomalous rainfall episodes.

\textbf{(58) \emph{Longest consecutive wet-hour run --- }}length of the
longest uninterrupted sequence of hours with \emph{Total precipitation}
exceeding 0.1\,mm\,h\ensuremath{^{-1}}, encoding temporal persistence and clustering of
rainfall.

\textbf{(59) \emph{Hourly precipitation concentration (Gini coefficient)
--- }}Gini coefficient of the hourly \emph{Total precipitation}
distribution within the bin, measuring the degree of concentration or
peakedness (0\,=\,uniform distribution; values approaching 1 indicate
concentration in a small number of hours).

\textbf{(60) \emph{Extreme-rain hour count --- }}number of hours with
\emph{Total precipitation} exceeding 10\,mm\,h\ensuremath{^{-1}}, capturing the occurrence
frequency of extreme-intensity rainfall relevant to surface ponding and
rapid runoff generation.

Together, the 13 Tier 2 features complement the Tier 1 scalar
statistics by encoding the dynamic, cross-variable, and distributional
properties of meteorological forcing that scalar summaries cannot
represent. Process 4 is particularly targeted at flood-relevant signals:
the extreme-event statistics directly characterize rainfall conditions
known to precede flash-flood initiation.

\hypertarget{appendix-d.-detailed-cross-window-evolution-descriptors-tier-3}{%
\section{Detailed Cross-Window Evolution Descriptors (Tier 3)}\label{appendix-d.-detailed-cross-window-evolution-descriptors-tier-3}}

This appendix documents Tier 3, which occupies the 10 dimensions at
window indices 60--69 (0-based) within each 24-hour bin. Tier 3 encodes
the between-bin memory, trajectory, and acceleration information that
scalar within-bin summaries cannot express. The six 24-hour bins are
indexed k = 0, \ldots, 5 from the farthest bin (k\,=\,0, earliest
antecedent period) to the nearest bin (k\,=\,5, immediately preceding the
flood peak). All 10 Tier-3 features are deterministic functions of the
per-bin Tier-1 vectors of bins 0 through k and do not invoke any learned
model. They are generated by four processes: antecedent/cumulative state
(P1, 3 scalars), inter-bin first differences (P2, 3 scalars), multi-bin
linear trends and acceleration (P3, 3 scalars), and a normalized
lookback position index (P4, 1 scalar).

\textbf{Process 1: Antecedent and Cumulative State (3 features, indices
60--62)}

Process 1 captures accumulated hydrometeorological state from the start
of the 144-hour lookback through the current bin. Antecedent soil
moisture, which governs infiltration capacity and the runoff fraction,
is determined largely by the precipitation history preceding the
triggering event. Three scalars are computed.

\textbf{(61) \emph{Cumulative Precipitation}}

The first descriptor is the \emph{Cumulative precipitation} up to the
current bin:

\[E_{i}^{(1)}\  = \ \sum_{j = 1}^{i}P_{j}\]

where \emph{P}\ensuremath{_{j}} denotes the \emph{Total precipitation} (Tier-1 index 0)
within bin \emph{W}\ensuremath{_{j}}. This variable captures total storm water input
accumulated over the antecedent period, serving as a direct indicator of
watershed wetting and catchment loading.

\textbf{(62) \emph{Antecedent Precipitation Index }}

To represent hydrological memory with temporal decay, the
\emph{Antecedent Precipitation Index} is computed as:

\[E_{i}^{(2)}\  = \ \sum_{j = 1}^{i}{\lambda^{i - j}{\ P}_{j}}\]

where \ensuremath{\lambda}\,=\,0.85 is the decay coefficient. This formulation, following the
classical API concept in operational hydrology, assigns greater weight
to recent rainfall while preserving older events' contributions with
exponentially diminishing effect.

\textbf{(63) \emph{Cumulative Net Water Input }}

The third descriptor aggregates \emph{Net water input} (P−ET\ensuremath{_{0}}) over
preceding bins:

\[E_{i}^{(3)}\  = \ \sum_{j = 1}^{i}\left( P_{j}{\  - \ ET}_{j} \right)\]

where ET\emph{\ensuremath{_{j}}} denotes total \emph{Total reference evapotranspiration
FAO-56 ET\ensuremath{_{0}}} within bin \emph{W}\ensuremath{_{j}}. This descriptor better approximates
the effective atmospheric water forcing than cumulative precipitation
alone, by accounting for evapotranspiration losses.

\textbf{Process 2: Inter-Bin First Differences (3 features, indices
63--65)}

Process 2 encodes the bin-to-bin rate of change of key meteorological
variables by computing the first difference between the current bin
\emph{i} and the immediately preceding bin \emph{i−1}. These signed
differences allow the model to detect intensification or attenuation of
flood-generating weather systems across consecutive 24-hour periods.

\textbf{(64) \emph{Inter-Bin Precipitation Change }}

The fourth descriptor captures changes in \emph{Total precipitation}:

\[\left\{ \begin{array}{r}
P_{i}{\  - \ P}_{i - 1},\quad i\  > \ 1 \\
0,\quad i\  = \ 1
\end{array} \right.\ \]

Positive values indicate intensifying precipitation; negative values
indicate recession. For the first bin (\emph{i\,=\,1}), this feature is
set to zero.

\textbf{(65) \emph{Inter-Bin Change in Sea-Level Pressure Trend }}

Let \emph{T}\ensuremath{_{i}}\ensuremath{^{(}}\ensuremath{^{p}}\ensuremath{^{)}} denote the \emph{Linear trend of sea-level pressure}
(Tier-1 index 21) in bin \emph{W}\ensuremath{_{i}}. The fifth descriptor is:

\[\left\{ \begin{array}{r}
T_{i}^{(p)}{\  - \ T}_{i - 1}^{(p)},\quad i\  > \ 1 \\
0,\quad i\  = \ 1
\end{array} \right.\ \]

This variable captures acceleration or deceleration in synoptic pressure
evolution, distinguishing a steadily deepening low (constant negative
slope) from a rapidly deepening system (increasingly negative slope).

\textbf{(66) \emph{Inter-Bin Change in Mean Relative Humidity }}

Let \emph{H}\ensuremath{_{i}} denote the \emph{Mean 2-meter relative humidity} (Tier-1
index 15) in bin \emph{W}\ensuremath{_{i}}:

\[\left\{ \begin{array}{r}
H_{i}{\  - \ H}_{i - 1},\quad i\  > \ 1 \\
0,\quad i\  = \ 1
\end{array} \right.\ \]

Increasing humidity across consecutive bins indicates the approach of a
moisture-laden air mass and the pre-conditioning of the lower
troposphere for sustained precipitation.

\textbf{Process 3: Multi-Bin Trends and Precipitation Acceleration (3
features, indices 66--68)}

Process 3 estimates the overarching trajectory of precipitation and
synoptic pressure across all bins from 1 through \emph{i} by fitting
least-squares linear trends to the corresponding Tier-1 scalar
sequences. An additional second-difference scalar captures precipitation
acceleration.

\textbf{(67) \emph{Multi-Bin Linear Trend of Precipitation }}

The seventh descriptor is the least-squares slope of the sequence \{P\ensuremath{_{1}},
P\ensuremath{_{2}}, \ldots, P\ensuremath{_{i}}\} of \emph{Total precipitation} values:

\[E_{i}^{(7)}{\  = \ slope(P}_{1}{,\ P}_{2}{,\ \ldots,\ P}_{i})\]

A positive slope indicates multi-day intensification of rainfall; a
negative slope indicates attenuation. For \emph{i\,=\,1}, this feature is
set to zero by construction.

\textbf{(68) \emph{Multi-Bin Linear Trend of Mean Sea-Level Pressure }}

Let \emph{M}\ensuremath{_{i}} denote the \emph{Mean sea-level pressure} (Tier-1 index
19) in bin \emph{W}\ensuremath{_{i}}. The eighth descriptor is:

\[E_{i}^{(8)}\  = \ slope({\overline{M}}_{1},\ {\overline{M}}_{2},\ \ldots,\ {\overline{M}}_{i})\]

Sustained negative values indicate a prolonged cyclonic deepening
trajectory consistent with the approach of a major storm system.

\textbf{(69) \emph{Precipitation Acceleration }}

To capture second-order temporal changes, the \emph{Precipitation
acceleration} is defined as:

\[\left\{ \begin{array}{r}
\left( P_{i}{\  - \ P}_{i - 1} \right)\  - \ \left( P_{i - 1}{\  - \ P}_{i - 2} \right),\quad i\  \geq \ 3 \\
0,\quad i\  < \ 3
\end{array} \right.\ \]

Positive values indicate that rainfall is intensifying at an increasing
rate (convex trajectory); negative values indicate deceleration (concave
trajectory). For \emph{i\,\textless{}\,3}, this feature is set to zero.

\textbf{Process 4: Normalized Lookback Position (1 feature, index 69)}

Process 4 contributes a single positional scalar encoding the temporal
location of the current bin. This serves as an explicit inductive bias:
features from bins close to the flood peak carry different predictive
significance than those from earlier bins, and the position index allows
the model to learn this asymmetry without requiring separate model
instances per bin.

\textbf{(70) \emph{Normalized Lookback Position}}

The final descriptor is:

\[E_{i}^{(10)}\  = \ \frac{i\  - \ 1}{N\  - \ 1}\]

where \emph{N}\,=\,6 in our implementation. The value 0 corresponds to the
farthest bin (earliest antecedent period, 121--144\,h before the flood
peak) and 1 to the nearest bin (0--24\,h before the peak). This linear
encoding allows the model to condition on temporal proximity without
requiring positional embeddings.

\emph{\textbf{Summary of the Descriptor Set}}

The cross-window evolution vector for the \emph{i}-th bin is:

\[e_{i}\  = \ \left\lbrack E_{i}^{(1)},\ E_{i}^{(2)},\ E_{i}^{(3)},\ E_{i}^{(4)},\ E_{i}^{(5)},\ E_{i}^{(6)},\ E_{i}^{(7)},\ E_{i}^{(8)},\ E_{i}^{(9)},\ E_{i}^{(10)} \right\rbrack\]

These 10 descriptors are grouped into three conceptual categories:

\begin{quote}
• \emph{\textbf{Cumulative-memory descriptors:}} Cumulative
precipitation, Antecedent Precipitation Index, and Cumulative net water
input.

• \emph{\textbf{Transition descriptors:}} Inter-bin precipitation
change, Inter-bin change in sea-level pressure trend, and Inter-bin
change in mean relative humidity.

• \emph{\textbf{Evolution-shape descriptors:}} Multi-bin linear trend of
precipitation, Multi-bin linear trend of mean sea-level pressure,
Precipitation acceleration, and Normalized lookback position.
\end{quote}

\bibliography{refs}

\end{document}